\documentclass{article}
\usepackage[preprint]{conference}
\usepackage{style}

\usepackage{microtype}
\usepackage{booktabs}
\usepackage{hyperref}
\usepackage{url}
\usepackage{lineno}

\usepackage[capitalize,noabbrev]{cleveref}
\usepackage[textsize=tiny]{todonotes}
\usepackage{tabularx}   %
\usepackage{array}      %
\usepackage{ragged2e}   %

\title{\raisebox{-0.35\height}{\includegraphics[height=1.4cm]{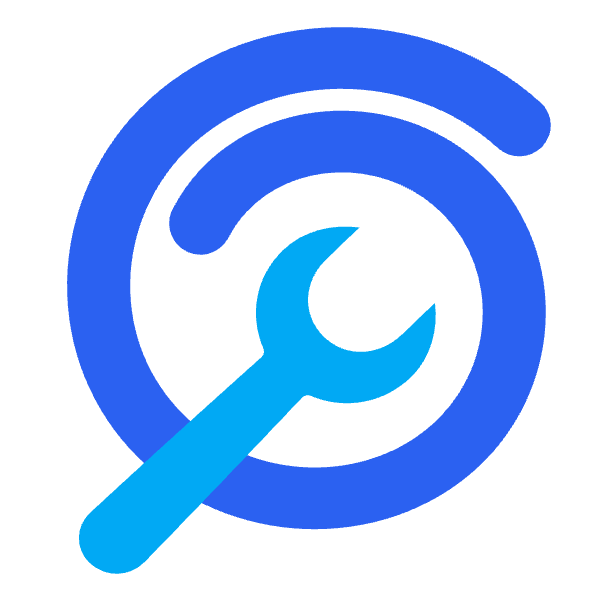}}\hspace{0.6em}\begin{tabular}[c]{@{}l@{}}\LARGE{The Finetuner's Fallacy}\\\Large{When to Pretrain with Your Finetuning Data}\end{tabular}}

\author{DatologyAI Team\thanks{See Contributions (\S~\ref{sec:contri}) for full author list.}}

\begin{document}

\maketitle

\begin{abstract}
\begin{adjustbox}{minipage=0.8\textwidth, center}
Real-world model deployments demand strong performance on narrow domains where data is often scarce. Typically, practitioners finetune models to specialize them, but this risks overfitting to the domain and forgetting general knowledge.
We study a simple strategy, \emph{specialized pretraining} (SPT), where a small domain dataset, typically reserved for finetuning, is repeated starting from pretraining as a fraction of the total tokens.
Across three specialized domains (ChemPile, MusicPile, and ProofPile), SPT improves domain performance and preserves general capabilities after finetuning compared to standard pretraining. In our experiments, SPT reduces the pretraining tokens needed to reach a given domain performance by up to 1.75×. These gains grow when the target domain is underrepresented in the pretraining corpus: on domains far from web text, a 1B SPT model outperforms a 3B standard pretrained model. Beyond these empirical gains, we derive \emph{overfitting scaling laws} to guide practitioners in selecting the optimal domain-data repetition for a given pretraining compute budget.
Our observations reveal the \emph{finetuner's fallacy}: while finetuning  may appear to be the cheapest path to domain adaptation, introducing specialized domain data during pretraining stretches its utility. 
SPT yields better specialized domain performance (via reduced overfitting across repeated exposures) and better general domain performance (via reduced forgetting during finetuning), ultimately achieving stronger results with fewer parameters and less total compute when amortized over inference.
To get the most out of domain data, incorporate it as early in training as possible.
\end{adjustbox}
\end{abstract}

\begin{figure*}[h]
\centering
\begin{minipage}{0.47\linewidth}
\vspace{-1.5em}
\includegraphics[width=\linewidth]{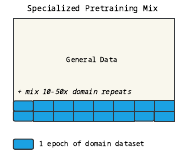}
\end{minipage}
\begin{minipage}{0.5\linewidth}
\vspace{0pt}
\includegraphics[width=\linewidth]{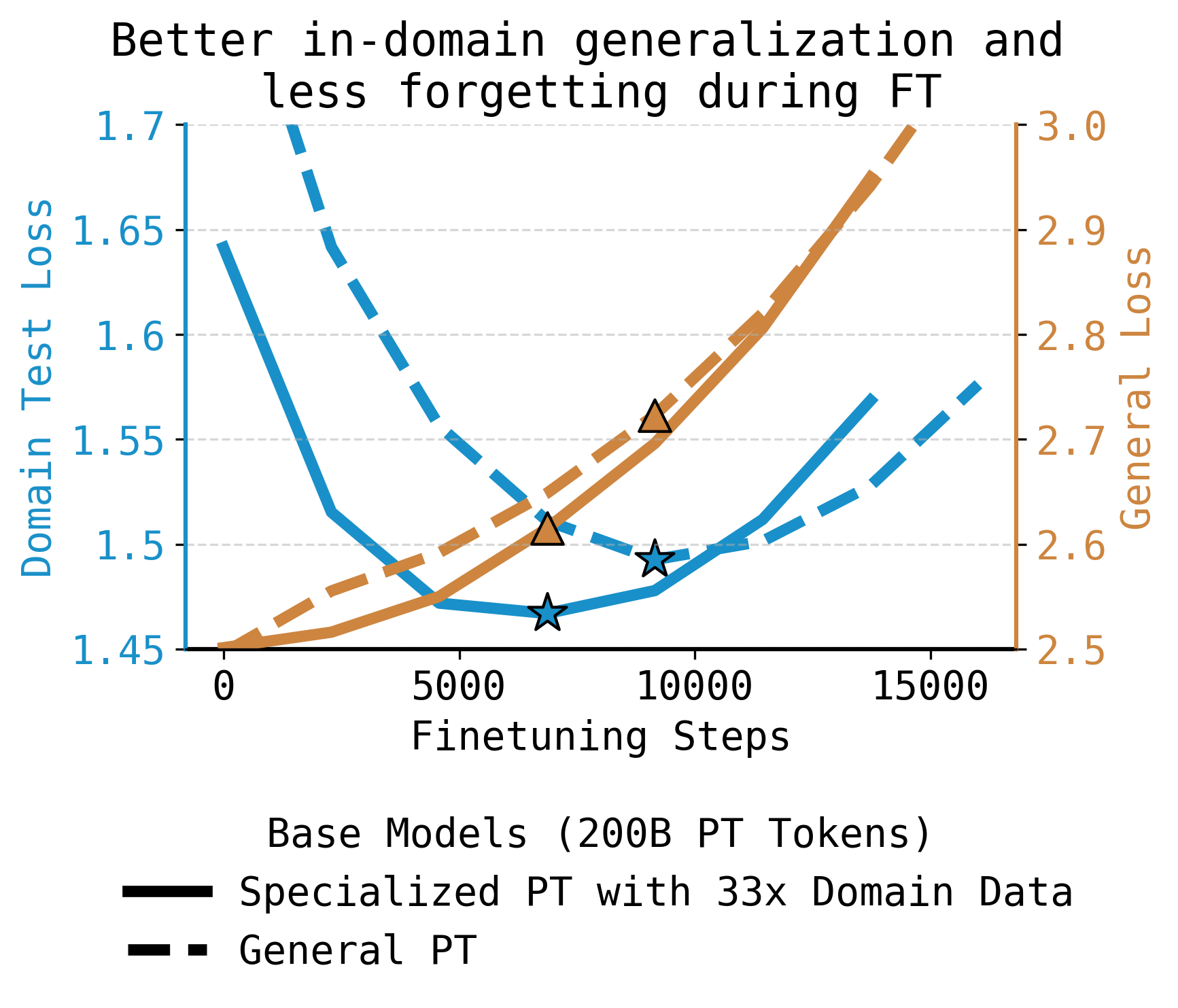}
\end{minipage}
\caption{Specialized pretraining (SPT) mixes the finetuning dataset into pretraining as a small fraction of tokens, repeating it 10–50× over the course of training. Compared to general pretraining (dashed), SPT (solid) achieves lower domain test loss (blue) and less forgetting of general knowledge (gold) throughout finetuning. For narrow domains, these gains can overcome differences in model scale.
\label{fig:first_page}}
\end{figure*}

\newpage

\section{Introduction}
Consider an organization with proprietary data such as support conversations, legal filings, or clinical notes, that wants to train a domain-specialized model. The conventional recipe is straightforward: start from a strong open-weights model pretrained on web-scale data, then finetune it on the proprietary dataset. Because this data is private and absent from public corpora, finetuning is treated as the natural mechanism for injecting missing domain knowledge. More broadly, modern training pipelines often treat pretraining and finetuning as disjoint phases: first learn general knowledge at scale, then specialize using a small curated dataset. The success of instruction tuning, RLHF, and parameter-efficient finetuning has further reinforced this view \citep{ouyang2022instructgpt,wei2021flan,hu2022lora}.

Growing evidence across several settings suggests that data encountered during
pretraining shapes model behavior more durably than that introduced
later: incorporating reasoning data into pretraining outperforms introducing it only
during fine-tuning~\citep{akter2025frontloading,hatamizadeh2025rlpreinforcementpretrainingobjective},
unsafe behaviors learned during pretraining are harder to remove via
post-training~\citep{maini2025safety,sam2025whensafety}, and cross-language transfer
during pretraining improves performance for low-resource
languages~\citep{longpre2025atlasadaptivetransferscaling}. Yet the question of
when to introduce domain-specific data, and whether it should be mixed
into pretraining rather than reserved for finetuning, remains largely unexplored.

In this work, we question whether reserving all domain-specific data for the final stage
of training is optimal. When the target domain is poorly represented in the pretraining
corpus, introducing domain data only during finetuning may require large representational
updates, leading to weaker generalization and greater forgetting of general knowledge.
We study a simple alternative: interleave the domain dataset throughout pretraining as a
small fraction of the training tokens (often repeating it up to 50 times), and then
finetune on the same data as usual (Figure~\ref{fig:first_page}). We refer to this
strategy as \emph{specialized pretraining} (SPT). We observe that interleaving domain tokens with general data allows the model to tolerate far more repetitions before overfitting than it would during finetuning.

\begin{figure}[!b]
    \centering    
    \includegraphics[width=\textwidth]{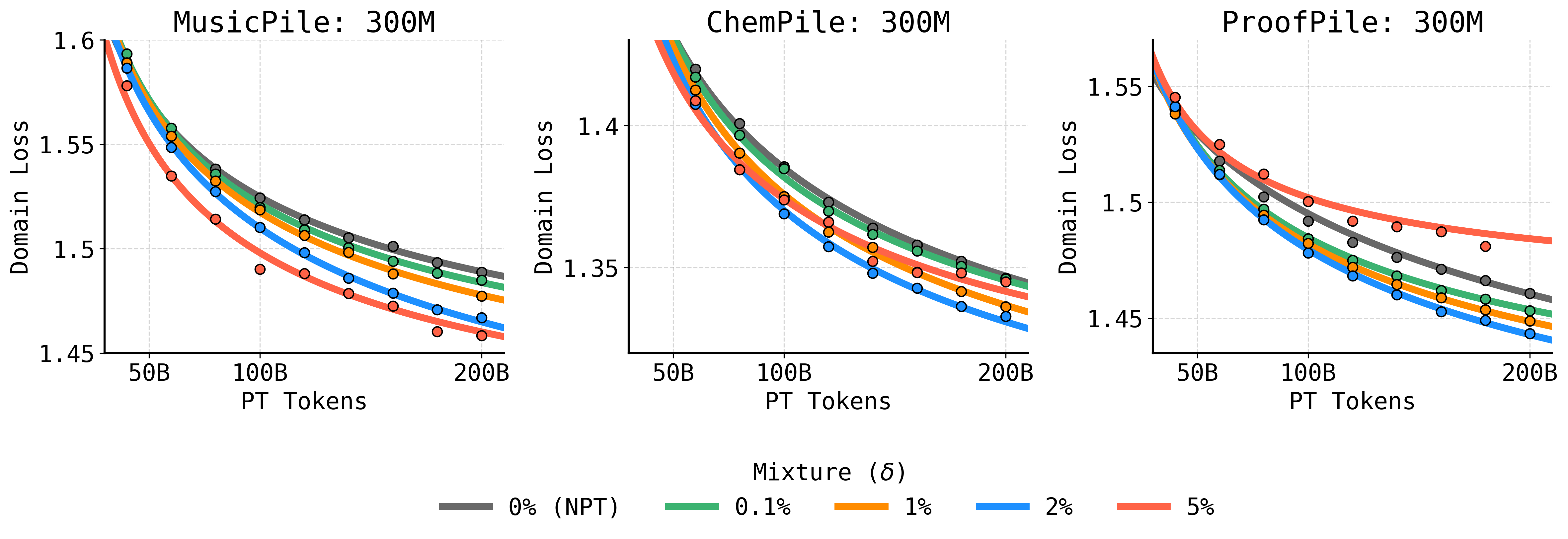}
    \caption{\textbf{Specialized pretraining (\texttt{SPT}) outperforms finetuning-only
    across domains.} We pretrain models with a small fraction ($\delta$) of domain-specific
    tokens mixed into general web data, then finetune on the domain dataset. We plot the
    best post-finetuning domain loss across pretraining budgets for MusicPile, ChemPile,
    and ProofPile (300M tokens each). Even small domain mixtures ($\delta = 1$--$5\%$,
    blue curves) consistently outperform pretraining on general data alone ($\delta = 0\%$,
    gray) at all token scales.}
    \label{fig:main-results}
\end{figure}

Across chemistry, symbolic music, and mathematical proofs, SPT consistently improves
post-finetuning performance (\S~\ref{sec:spt}). Relative to standard pretraining
followed by finetuning, models trained with SPT achieve lower domain test loss, retain
general pretraining knowledge more effectively during finetuning, and perform better on
downstream tasks (Figure~\ref{fig:main-results}). Comparing across pretraining scales,
SPT also requires substantially less pretraining compute to reach the same
post-finetuning domain loss. 
Even replay~\citep{parmar2024cpt}, a common strategy that reintroduces general data during continued pretraining, is not a substitute for early domain exposure: SPT's advantage persists across all replay settings (\S~\ref{sec:replay}). On domains far from web text, a 1B model trained with SPT outperforms a 3B model trained without domain data during pretraining. 
SPT also reduces the pretraining tokens needed to reach a given domain loss by up to 1.75$\times$ (Figure~\ref{fig:main-results}), and these loss improvements translate to downstream task accuracy: at 200B pretraining tokens, SPT improves MATH accuracy by up to 6 percentage points and MusicTheoryBench by up to 4 percentage points over the finetuning-only baseline (Figure~\ref{fig:downstream}).

We further characterize when SPT is most effective. First, SPT helps most when the \textbf{target domain is underrepresented} in the pretraining corpus, as shown both in a controlled multilingual overlap study and across naturally occurring domain shifts~(\S~\ref{sec:domain-similarity}). Second, its benefit depends on the \textbf{size of the domain dataset}: for sufficiently large domain corpora, even repeating domain data from the beginning of pretraining is beneficial, while in more data-constrained settings, introducing domain data later in pretraining is preferable (\S~\ref{subsec:dataset-size}). 
Third, the \textbf{pretraining budget} determines the optimal mixture fraction: larger 
domain fractions help at shorter training horizons, while smaller fractions become 
preferable as training lengthens to avoid overfitting from excessive repetition 
(\S~\ref{subsec:compute-budget}).

Finally, we derive \textbf{overfitting scaling laws} to model the total overfitting incurred by specialized pretraining and finetuning for different mixture fractions (\S~\ref{sec:scaling-laws}), which include new expressions for modeling the effects of repeated data. Specifically, the test loss decomposes as the sum of two parts: the training loss w.r.t. pretraining tokens is modeled as a power law with a negative exponent, while the growing train-test gap is modeled as a power law with a positive exponent. This allows us to predict the optimal domain-data fraction for a given compute budget and forecast when aggressive mixing begins to hurt test loss, without running the full training sweep. In practice, this means practitioners can select the right SPT configuration from a small number of pilot runs rather than exhaustive search.

\begin{figure*}[t]
    \centering
    \includegraphics[width=0.85\textwidth]{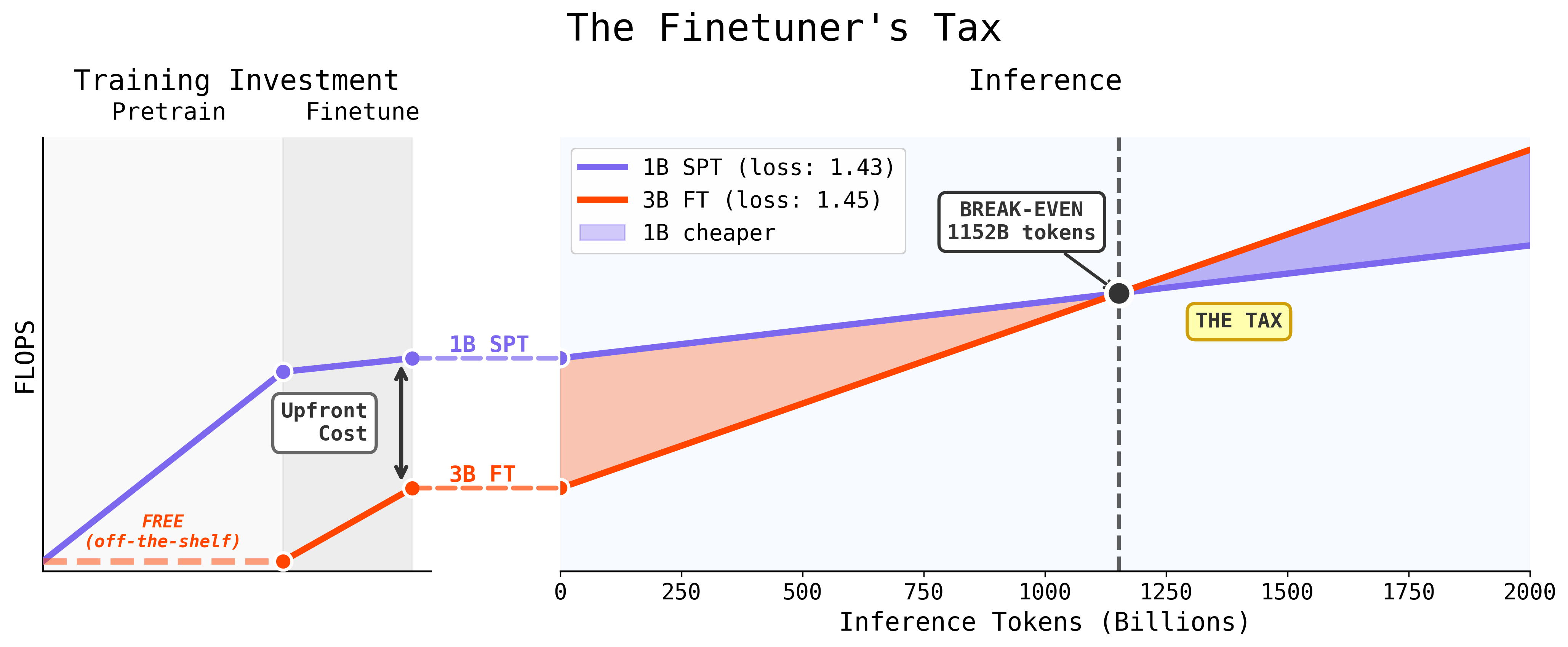}
    \caption{\textbf{The finetuner's tax.} Training a 1B model with specialized
    pretraining (\texttt{SPT}) costs more upfront than finetuning a 3B model on
    domain data alone, but the $3\times$ smaller model is cheaper to serve. The
    break-even point arrives after approximately 1 trillion inference tokens, after which
    \texttt{SPT} saves both compute and money while often delivering comparable or better performance.}
    \label{fig:finetuners_tax}
\end{figure*}

Broadly, our results expose \emph{the finetuner's fallacy}. Finetuning a large off-the-shelf model is commonly assumed to be the cheapest path to domain adaptation, since it avoids the cost of pretraining. 
This intuition is often misleading. Those who rely on finetuning alone need a larger model to match domain performance, and amortized over inference, pretraining a smaller model with domain data is cheaper (Figure~\ref{fig:finetuners_tax}). 
Early integration of domain data during pretraining is one way to reduce this cost, and is compatible with other data curation strategies like synthetic augmentation. 
As the cost of pretraining continues to fall, the case for early integration only strengthens. To get the most out of domain data, incorporate it as early in training as possible.

\begin{figure*}[b]
    \centering
    \includegraphics[width=\linewidth]{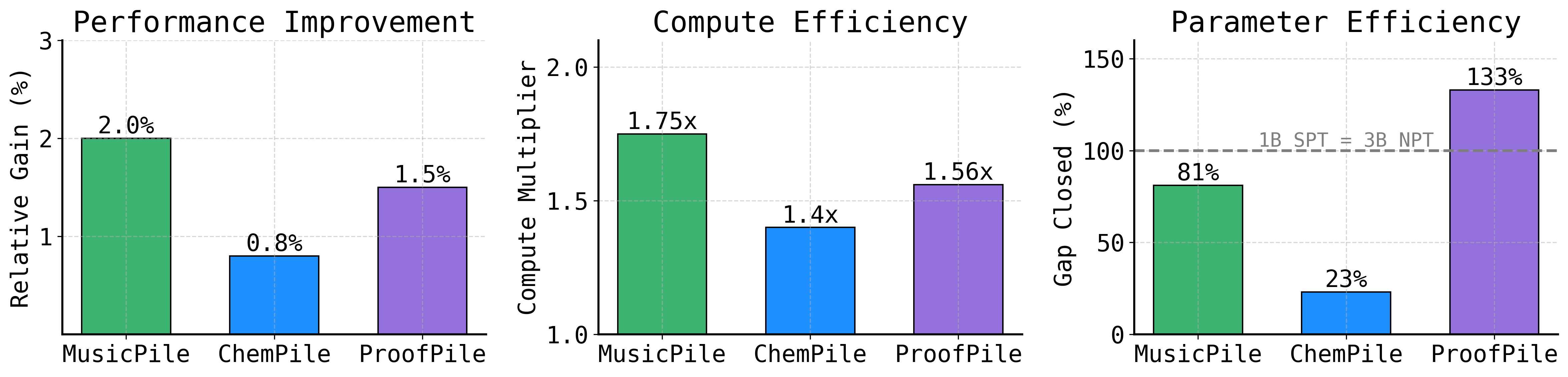}
    \caption{\textbf{Specialized pretraining (\texttt{SPT}) is more effective than scaling
tokens or parameters.} We compare models that include domain data during pretraining
(\texttt{SPT$\to$FT}) against models pretrained only on general web data
(\texttt{NPT$\to$FT}). \emph{Left:} Relative
gain of \texttt{SPT$\to$FT} over \texttt{NPT$\to$FT}. \emph{Center:} Compute
multiplier showing how much faster \texttt{SPT} reaches \texttt{NPT}'s best
performance. \emph{Right:} Percentage of the 1B vs.\ 3B parameter performance gap
closed by \texttt{SPT}. Values above 100\% indicate the 1B \texttt{SPT}
outperforms 3B \texttt{NPT}. \texttt{SPT} consistently improves model quality, training speed, and parameter efficiency (i.e. \texttt{SPT} is a pareto improvement in training efficiency) across all three examined domains.}
\label{fig:rgain-compute}
\end{figure*}

\section{Specialized Pretraining Drives Domain Specific Capabilities}
\label{sec:spt}
We study whether mixing domain-specific data into pretraining improves the model that
results after finetuning. We refer to standard pretraining on general web data as
\emph{naive pretraining} (\texttt{NPT}), and to pretraining that includes a small
fraction of domain data as \emph{specialized pretraining} (\texttt{SPT}). Both are
followed by finetuning (\texttt{FT}) on the domain dataset. We compare the two
resulting pipelines, \texttt{NPT$\to$FT} and \texttt{SPT$\to$FT}, across three
specialized domains.

\subsection{Notation and Experimental Setup}
\label{subsec:setup-notation}

\paragraph{Specialized Pretraining} Let $\delta \in [0,1]$ denote the fraction of pretraining tokens drawn from the domain-specific dataset, with the remaining $1-\delta$ fraction drawn from general web data (e.g., $\delta=0.02$ corresponds to a 2\% domain token mixture). Note that $\delta=0$ corresponds to naive pretraining. Since domain-specific datasets are typically much smaller than the total pretraining budget, domain examples are repeated as necessary during pretraining. Given $T$ pretraining tokens, the total epochs of domain-specific data $\mathcal{D}_{\text{dom}}$ seen is $E = (T \cdot \delta)/|\mathcal{D}_{\text{dom}}|.$ To measure generalization under this heavy repetition, we hold out a fixed test split from each domain dataset. This split is never included in training, and all domain losses reported in this paper are evaluated on it.

\paragraph{OLMo Sandbox} We use a controlled pretraining environment based on OLMo-1B trained on the Dolma corpus. We introduce three specialized domains of equal size, MusicPile \citep{musicpile}, ChemPile \citep{chempile}, and ProofPile \citep{proofpile}, each containing roughly 300M tokens. We pretrain for 200B tokens with $\delta\in\{0, 0.1\%, 1\%, 2\%, 5\%\}$ fraction drawn from the domain-specific dataset and the remainder $(1-\delta)$ drawn from Dolma. Samples are repeated to satisfy mixture constraints e.g. 5$\%$ \texttt{SPT} repeats the domain-specific data roughly $33\times$ over the course of pretraining. We match OLMo-1B pretraining settings to the publicly documented configuration where possible, including the optimizer, cosine learning rate schedule, and batch size. We report all hyperparameters in Appendix~\ref{app:hyperparameters}.

We will also compare SPT and NPT across model sizes, for which we create 300M, 600M, and 3B variants of the OLMo-1B architecture by adjusting the model depth. For the 3B variant, we double the number of layers in the 1B model. For smaller 300M and 600M variants, we halve the number of layers and reduce the dimension of the MLP representations and hidden representations. We train all models using the same pretraining recipe.

\paragraph{Finetuning} During finetuning, training proceeds exclusively on the
domain-specific dataset with a WSD learning rate schedule. To ensure a fair comparison
of post-finetuning performance between \texttt{SPT} and \texttt{NPT} models, we impose
no restriction on the number of dataset repetitions. Instead, we apply early stopping:
the data is repeated as long as test loss decreases. This ensures that any loss
improvement observed with \texttt{SPT} is not a result of additional passes over
the dataset. Furthermore, we tune warmup steps and the
learning rate by grid search. We report the  lowest domain test loss across all
finetuning configurations.

\subsection{Experimental Results}
\label{subsec:results}
We compare two training paradigms. In the first, \texttt{NPT$\to$FT}, we pretrain over Dolma for 200B tokens ($\delta=0$) then finetune. In the second, \texttt{SPT$\to$FT}, we pretrain on repeats of domain-specific data mixed with Dolma for 200B tokens before finetuning.

\paragraph{Which paradigm achieves lower domain test loss?} Let
$L^{\texttt{NPT}\to\texttt{FT}}$ and $L^{\texttt{SPT}(\delta)\to\texttt{FT}}$ denote
the best domain test loss for each paradigm. We quantify improvement by \emph{relative
gain}:
\begin{align*}
\mathcal{R}_{\text{gain}}(\delta) \;=\; 100 \cdot
\frac{L^{\texttt{NPT}\to\texttt{FT}} -
L^{\texttt{SPT}(\delta)\to\texttt{FT}}}{L^{\texttt{NPT}\to\texttt{FT}}},
\end{align*}
\texttt{SPT} consistently helps across all three domains
(Figure~\ref{fig:rgain-compute}, left). MusicPile, a corpus of symbolic music notation
that bears little resemblance to web text, sees the largest relative gain at 2.0\%.
ProofPile, containing formal mathematical proofs, follows with 1.5\%. ChemPile shows a
more modest 0.8\% improvement, which we attribute to chemistry text being closer to the
Dolma distribution. We return to the impact of distributional overlap in
\S~\ref{sec:domain-similarity}.

\begin{figure}[t]
    \centering
    \begin{subfigure}[t]{0.5\linewidth}
    \centering
    \vspace{0pt}
    \includegraphics[width=\linewidth]{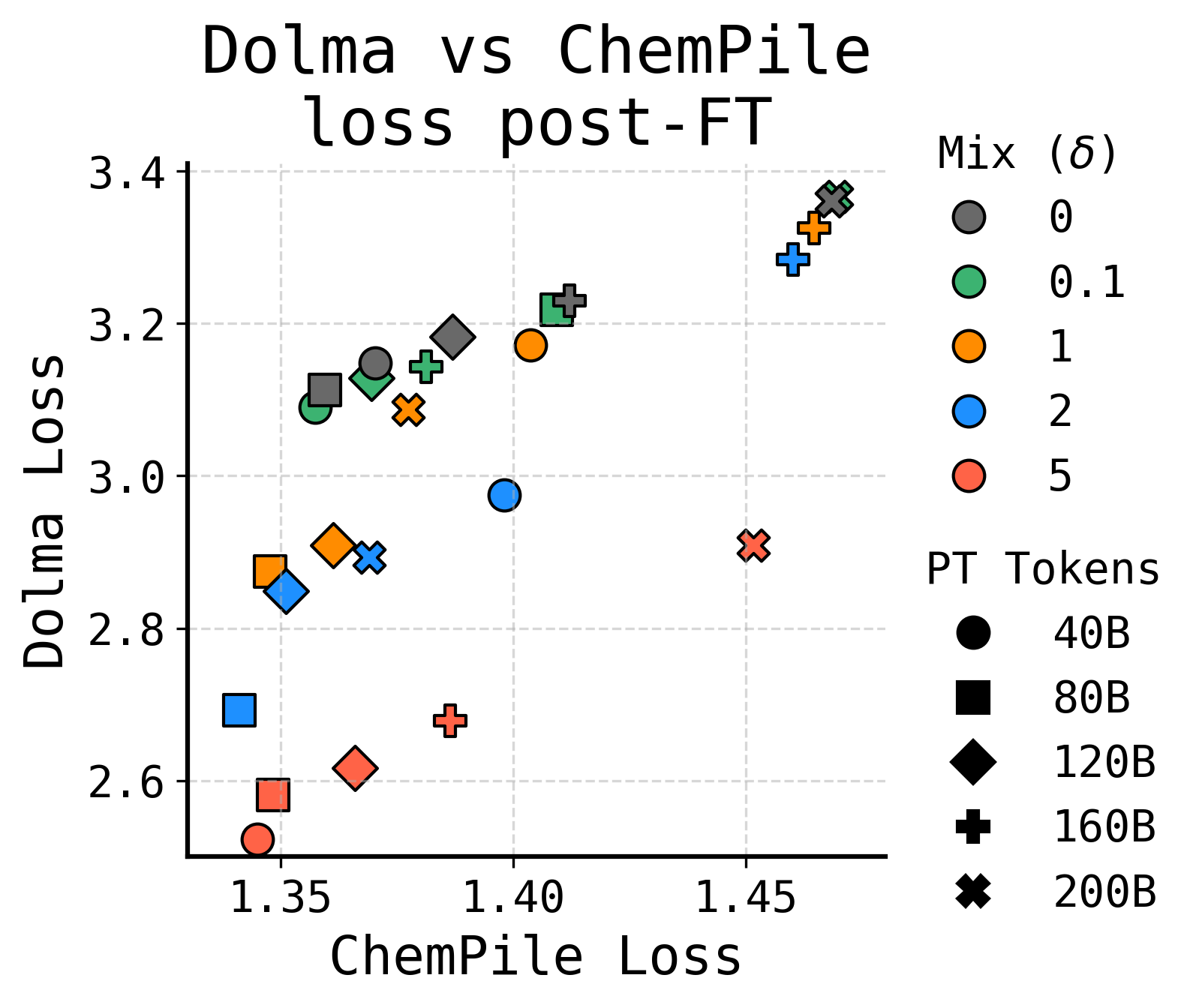}
    \caption{\label{subfig:forgetting}} 
    \end{subfigure} 
    \hfill
    \begin{subfigure}[t]{0.44\linewidth}
    \centering
    \vspace{0pt}
    \includegraphics[width=0.51\linewidth]{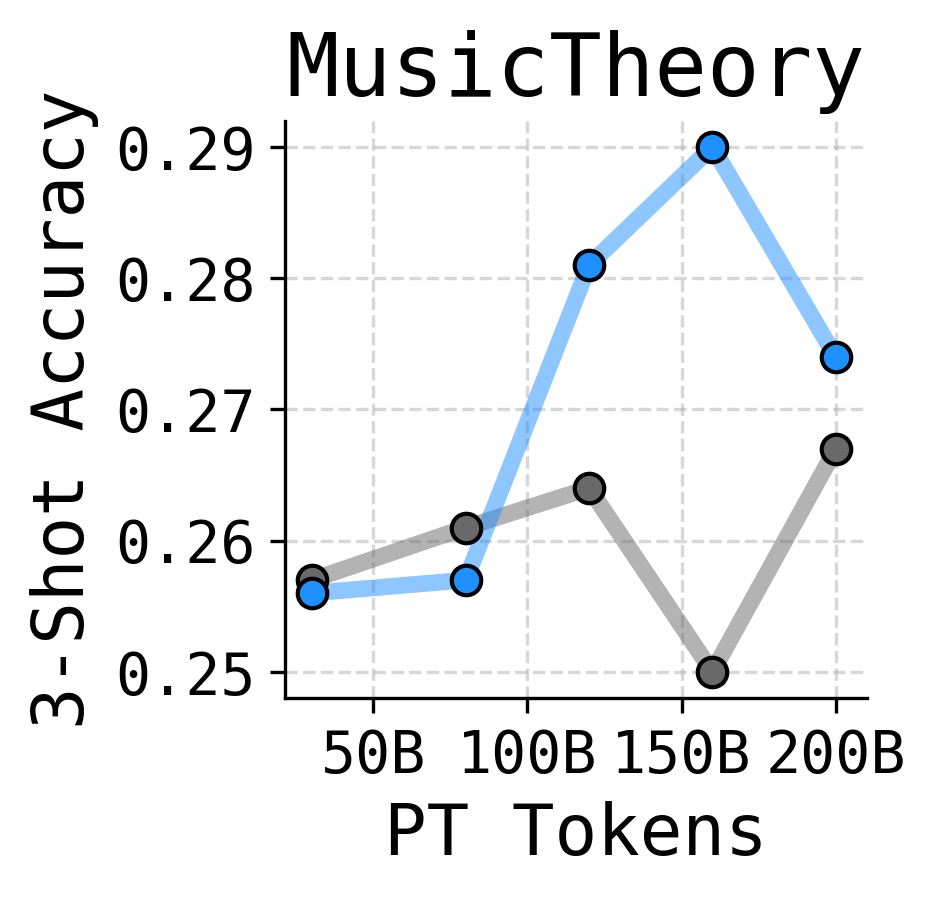}
    \newline
    \includegraphics[width=0.47\linewidth]{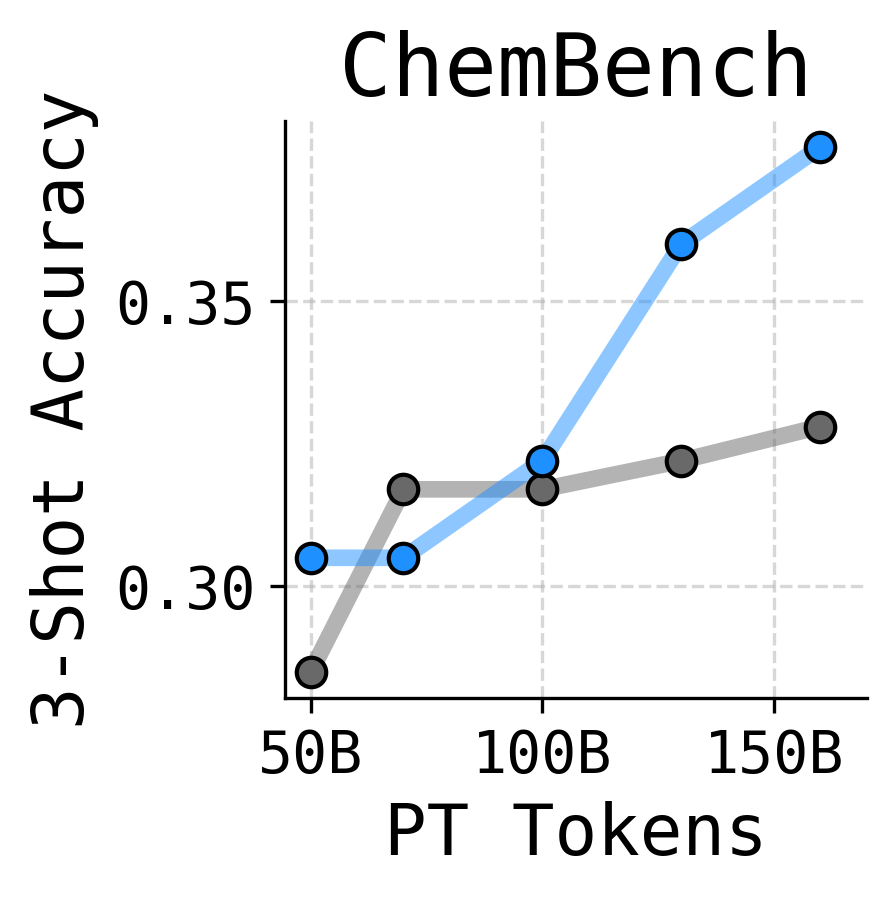}
    \includegraphics[width=0.51\linewidth]{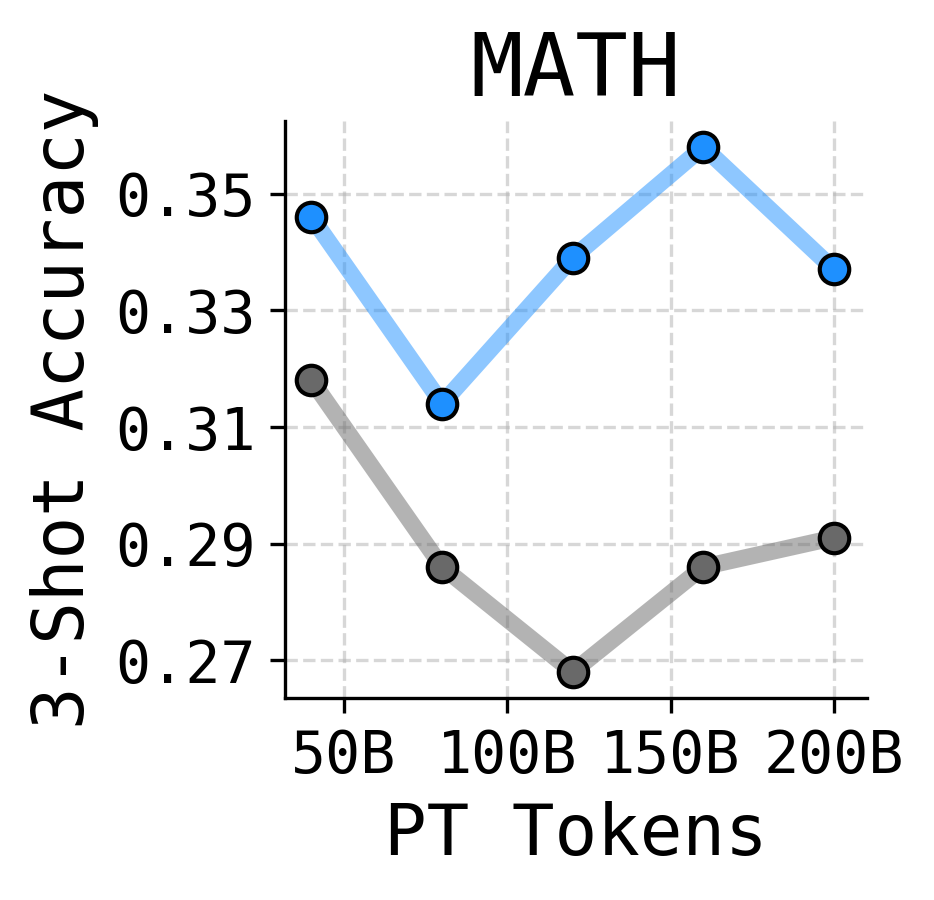}
    \caption{\label{subfig:downstream_tasks}}
    \end{subfigure}
    \caption{\textbf{\texttt{SPT} reduces forgetting and improves downstream task
    performance.} (a)~For ChemPile, we plot Dolma loss (general knowledge) against
    domain loss for the best post-finetuning checkpoint at each pretraining budget (40B
    to 200B tokens) and mixture percentage $\delta$. Larger \texttt{SPT} mixtures
    achieve lower domain loss \emph{and} lower general loss, indicating less catastrophic
    forgetting. (b)~We compare \texttt{NPT} (gray) and 2\% \texttt{SPT} (blue) on
    downstream tasks matched to each domain: MusicTheoryBench for MusicPile, ChemBench General Chemistry subset for ChemPile, and MATH for ProofPile. All tasks are evaluated in
    4-choice MCQA format. For each pretraining budget, we report the best accuracy across
    finetuning runs. \texttt{SPT} outperforms \texttt{NPT} across most settings.}
    \label{fig:downstream}
\end{figure}

\paragraph{How much earlier in pretraining does SPT reach a given domain test loss?} We
define the \emph{compute multiplier} as the factor by which \texttt{SPT} reduces the
number of pretraining tokens needed to reach a given post-finetuning domain test loss,
compared to \texttt{NPT}. The efficiency gains are substantial
(Figure~\ref{fig:rgain-compute}, center). On MusicPile, \texttt{SPT} matches
\texttt{NPT}'s best performance at 200B pretraining tokens using $1.75{\times}$ fewer tokens. ProofPile shows a
similar $1.56{\times}$ advantage. Even ChemPile, despite its smaller performance gap,
achieves a $1.40{\times}$ multiplier. 

\paragraph{Can SPT compensate for model size?} If a smaller model trained with
\texttt{SPT} could match or exceed the performance of a larger \texttt{NPT} model, the
practical implications would be substantial. To test this, we pretrain a 3B model on the
same 200B Dolma tokens and measure how much of the gap between the 1B and 3B
\texttt{NPT} models is closed by the 1B \texttt{SPT} model. The results vary by domain
(Figure~\ref{fig:rgain-compute}, right). On ProofPile, the furthest domain from web
text, the 1B \texttt{SPT} model closes 133\% of the gap, meaning it surpasses the 3B
model entirely. MusicPile closes 81\% of the gap, nearly matching the larger model.
ChemPile shows more modest gains at 23\%, consistent with its smaller relative gain
(\S~\ref{subsect:corr_anal}).

Taken together, \texttt{SPT} delivers better domain performance, faster convergence, and stronger parameter efficiency across all three domains, with no observed tradeoff between these axes.

\subsection{SPT Learns More and Forgets Less}
\label{subsec:downstream}
In addition to lower domain loss, \texttt{SPT} reduces forgetting of general knowledge
during finetuning. Although \texttt{SPT} allocates a small fraction of pretraining
tokens to domain data, this has minimal impact on Dolma loss during pretraining: the
\texttt{NPT} and \texttt{SPT} runs achieve comparable general loss after 200B tokens (Appendix \ref{app:general_loss}).
The difference emerges during finetuning. \texttt{NPT} models start with higher domain
loss and therefore require more aggressive optimization to close the gap, which drives up
general loss. \texttt{SPT} models start from a lower domain loss, require less
adaptation, and consequently exhibit less forgetting (Figure~\ref{subfig:forgetting}).

The combination of lower domain loss and lower general loss correlates with improved
downstream task performance. We compare the best
post-finetuning performance of a 2\% \texttt{SPT} model against an \texttt{NPT}
baseline on symbolic music questions from MusicTheoryBench
\citep{musicpile}, ChemBench general chemistry subset \citep{Mirza2025}, and MATH in
MCQA style \citep{math_mcqa_2025}. Across all benchmarks and most pretraining budgets,
\texttt{SPT} outperforms \texttt{NPT} by several percentage points (Figure~\ref{subfig:downstream_tasks}).

\subsection{Specialized Pretraining Reduces Overfitting During Finetuning}
\label{subsec:overfitting}
Prior work on mixing data during pretraining typically studies regimes in which data is
abundant and rarely repeated; in these settings, data mixing primarily serves to
accelerate optimization. Our setting differs fundamentally: the finetuning datasets are
at least an order of magnitude smaller than the model size (300M tokens versus 1B
parameters), meaning that with reasonable repetitions, the models can overfit to the
training data. In fact, across domains, for both \texttt{SPT$\to$FT} and
\texttt{NPT$\to$FT}, the domain test loss begins to rise after roughly 5 epochs of
finetuning (Figure~\ref{fig:first_page}). This suggests that the benefit of specialized
pretraining is better understood as a regularization effect rather than an optimization
one.

\begin{figure}[h!]
  \centering
  \includegraphics[width=0.55\linewidth]{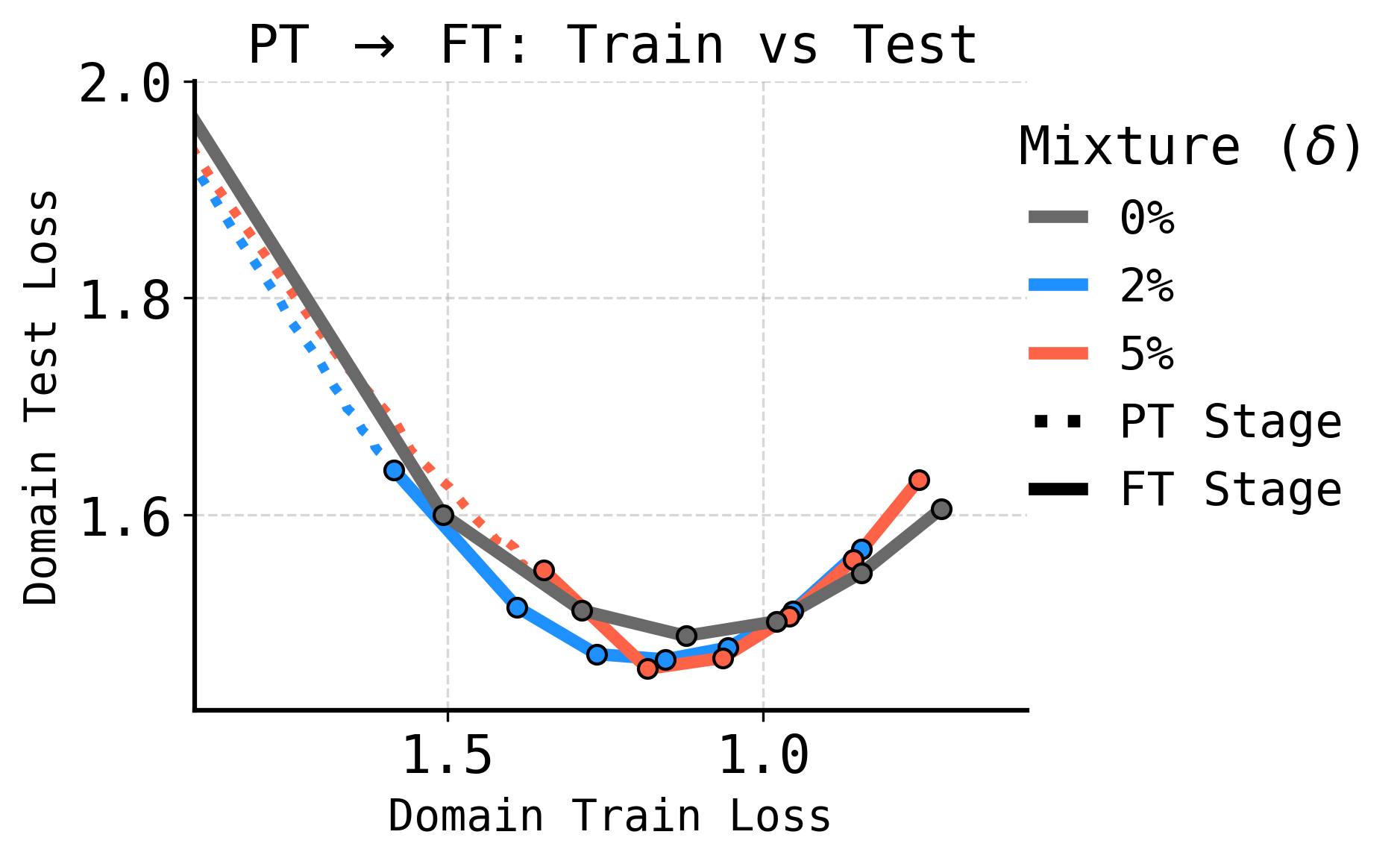}
  \caption{\textbf{\texttt{SPT} regularizes finetuning.} Domain training loss (x-axis)
  versus domain test loss (y-axis) across the pretraining and finetuning stages for
  MusicPile, with mixture fractions $\delta = 0\%, 2\%, 5\%$. Dotted lines show the
  pretraining stage; solid lines show finetuning. During finetuning, \texttt{SPT} models
  achieve lower test loss than \texttt{NPT} models at the same training loss, indicating
  that \texttt{SPT} reduces overfitting.
  Notably, the $\delta = 5\%$ model has already seen the domain data over $33\times$
  during pretraining, yet it still overfits less during finetuning than the \texttt{NPT}
  model seeing the same data for the first time. This regularization is a consequence of diffused exposure to specialized data during pretraining.}
  \label{fig:pt_to_ft_test_vs_train}
\end{figure}

To understand why \texttt{SPT} reduces overfitting, we compare how the domain train-test gap evolves across pretraining and finetuning (Figure~\ref{fig:pt_to_ft_test_vs_train}). The key insight is that overfitting to domain data is far more of a risk during finetuning than during pretraining. During pretraining, domain tokens make up only a small fraction of each batch, and the surrounding general data acts as a natural regularizer, preventing the model from memorizing the domain corpus even after tens of repetitions. 

During finetuning, this regularization effect is absent: the model trains exclusively on domain data and overfits rapidly. This is visible in Figure~\ref{fig:pt_to_ft_test_vs_train}: at the same domain training loss, comparing \texttt{SPT} at its initial pretrained checkpoint with \texttt{NPT} after early finetuning steps, the two models generalize comparably but as finetuning continues, the \texttt{NPT} model's train-test gap widens much faster. Because \texttt{SPT} models enter finetuning with a lower domain loss, they need less adaptation and exit finetuning before overfitting sets in.

\subsection{Key Takeaways}
\label{subsec:takeaways}
Overall, mixing domain-specific data into pretraining leads to a stronger model in terms
of \emph{both domain perplexity and downstream task performance}. Specialized pretrained
models generalize better during finetuning, which suggests that mixing the two datasets
together leads to qualitatively different learned representations. Furthermore, we find that retaining both general and domain performance simultaneously is a key
advantage of \texttt{SPT}. Because the domain data is introduced as a small fraction of
pretraining tokens alongside general web data, the model learns domain structure without
sacrificing broad coverage. During finetuning, this translates into less catastrophic
forgetting: \texttt{SPT} models start from a lower domain loss and therefore require
less aggressive adaptation, preserving general capabilities that \texttt{NPT} models
forfeit. This dual benefit is what makes \texttt{SPT} practically attractive, and what
underlies the finetuner's fallacy we characterize in the remainder of this paper.

\section{Factors governing the relative gain of specialized pretraining}
\label{sec:rgain-factors}
Section~\ref{sec:spt} established that \texttt{SPT} yields consistent gains over
\texttt{NPT} across diverse domains, improving domain test loss after finetuning while
also better retaining general knowledge. In this section, we characterize \emph{when} to
expect such gains. We identify several factors that impact the relative improvement,
including the similarity between pretraining and finetuning data,
the size of the domain dataset, and the available pretraining compute.

\subsection{Domain Similarity}
\label{sec:domain-similarity}

The degree of similarity between the domain-specific data and the general pretraining
corpus is expected to influence the efficacy of \texttt{SPT}. In the limiting case where
the pretraining and target distributions coincide, incorporating domain data during
pretraining introduces no additional signal and therefore should not improve downstream
performance. To examine this dependence, we analyze the effect of distributional overlap
through two complementary approaches: (i) a controlled study in which overlap is
systematically varied, and (ii) a cross-domain correlation analysis across naturally
occurring distribution shifts.

\subsubsection{A controlled study of distributional overlap}
To isolate the effect of distributional similarity with minimal confounding factors, we
consider an English$\to$Japanese translation setting. This task is well studied and
presents substantial linguistic divergence in script, morphology, and word order
\citep{nllbteam2022, fan2021beyond, liu2020mbart, carranza2026uberweb}. Standard practice
in machine translation is to pretrain on monolingual corpora from the source and target
languages and subsequently finetune on parallel data
\citep{xu2024paradigmshiftmachinetranslation, hangya-etal-2022-improving}. In our
setting, varying the proportion of Japanese relative to English monolingual text during
pretraining systematically alters the overlap between the pretraining distribution and
the downstream translation task. We compare \texttt{SPT$\to$FT} (where parallel
translation data is included during pretraining at mixture fraction $\delta$) against
\texttt{NPT$\to$FT} (monolingual pretraining only), while holding the total number of
pretraining tokens fixed.

\begin{figure}[hbt]
\centering
\includegraphics[width=0.65\linewidth]{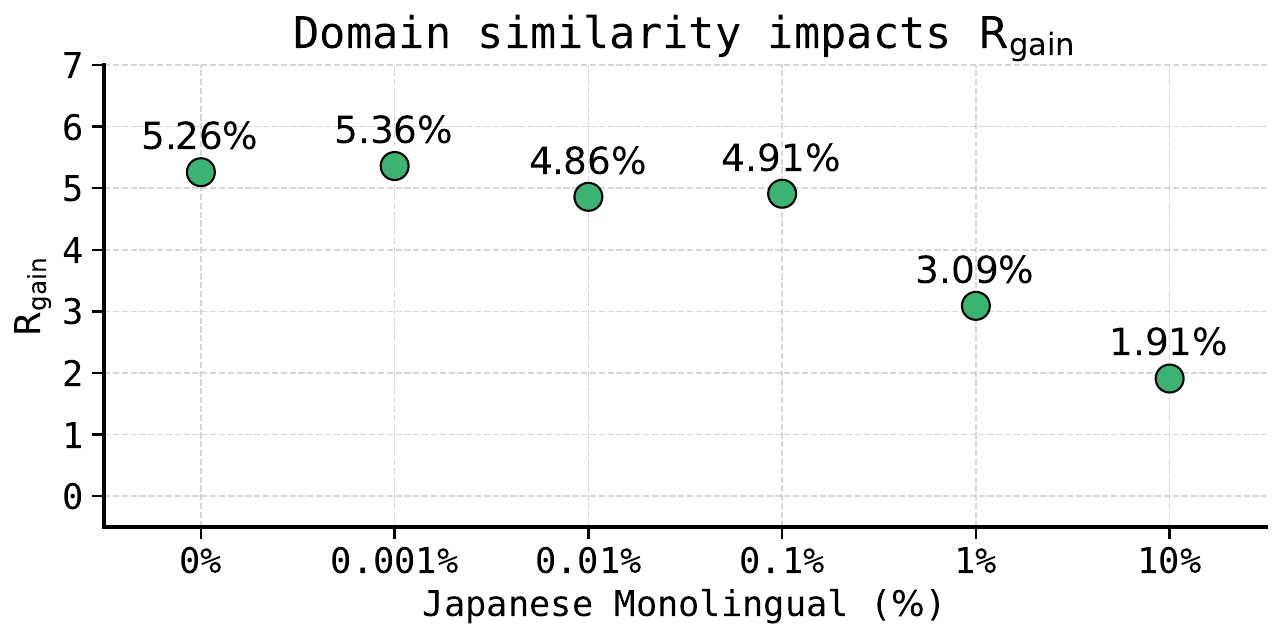}
  \caption{\textbf{Benefits of \texttt{SPT} increase as pretraining and finetuning
  domains diverge.} We vary the percentage of Japanese monolingual text in the
  pretraining mix for an English$\to$Japanese translation task, and plot
  $\mathcal{R}_{\text{gain}}$ of \texttt{SPT$\to$FT} over \texttt{NPT$\to$FT}. With
  less Japanese monolingual data (leftwards on x-axis), the distributional gap between
  pretraining and finetuning data grows, and the gain from \texttt{SPT} increases,
  plateauing at approximately 5\%.}
  \label{fig:domain-similarity-rgain}
\end{figure}

We pretrain 160M-parameter LLaMA models on 20B tokens of FineWeb2 English and Japanese
monolingual data \citep{penedo2025fineweb2pipelinescale}, together with 1B tokens of
parallel data from JParaCrawl v3.0 \citep{morishita-etal-2022-jparacrawl}. For
\texttt{SPT}, a fraction $\delta \in \{0\%, 0.1\%, 1\%, 5\%\}$ of pretraining tokens is
drawn from the parallel corpus, with the remaining $1-\delta$ drawn from a mixture of
English and Japanese monolingual data. We vary the proportion of Japanese monolingual
data within this mixture over $\{0\%, 0.001\%, 0.01\%, 0.1\%, 1\%, 10\%\}$. All models
are subsequently finetuned on the parallel corpus, and we select checkpoints based on
held-out validation loss.

\textbf{Results (Fig.~\ref{fig:domain-similarity-rgain}).} As the proportion of Japanese
monolingual data decreases from 10\% to 0.1\% in pretraining, we observe that
$\mathcal{R}_{\text{gain}}$ increases from approximately 2\% to 5\%. Increasing Japanese
monolingual content during pretraining reduces the distributional gap between pretraining
and finetuning (i.e., parallel translation) data, thereby diminishing the marginal
benefit of including parallel data early. These findings provide controlled evidence that
the relative gain from \texttt{SPT} is driven by distributional misalignment: when the
pretraining corpus already contains domain-relevant signal, incorporating domain data
during pretraining yields smaller improvements.

\subsubsection{Correlation Analysis Across Domains}
\label{subsect:corr_anal}

We next ask whether standard measures of distributional similarity between Dolma and
each specialized domain can predict $\mathcal{R}_{\text{gain}}$. We consider five
metrics: unigram, bigram, and trigram Jensen--Shannon divergence (JSD), MAUVE
\citep{pillutla2021mauve}, and the classifier two-sample test (C2ST)
\citep{lopezpaz2016c2st}. We also use a direct proxy: the domain loss of
\texttt{NPT} checkpoints after finetuning, since higher post-finetuning loss indicates
less overlap between the general corpus and the target domain.

In the controlled Japanese overlap sweep, all metrics correlate strongly with
$\mathcal{R}_{\text{gain}}$ ($|r| > 0.85$). However, when we measure distributional
overlap between Dolma and the three benchmark domains (MusicPile, ChemPile, ProofPile),
the metrics disagree on which domain is closest to Dolma. From
Section~\ref{subsec:results}, we know that $\mathcal{R}_{\text{gain}}$ is highest for
MusicPile (2.0\%), followed by ProofPile (1.5\%) and ChemPile (0.8\%). Of the metrics
we tested, only the post-finetuning domain loss, a direct measure of how well
\texttt{NPT} generalizes to each domain, correctly ranks all three domains (detailed analysis in Appendix~\ref{app:distribution-metrics}).

\begin{figure}[t]
    \centering
    \begin{minipage}[t]{0.48\linewidth}
    \vspace{0pt}
    \includegraphics[width=\linewidth]{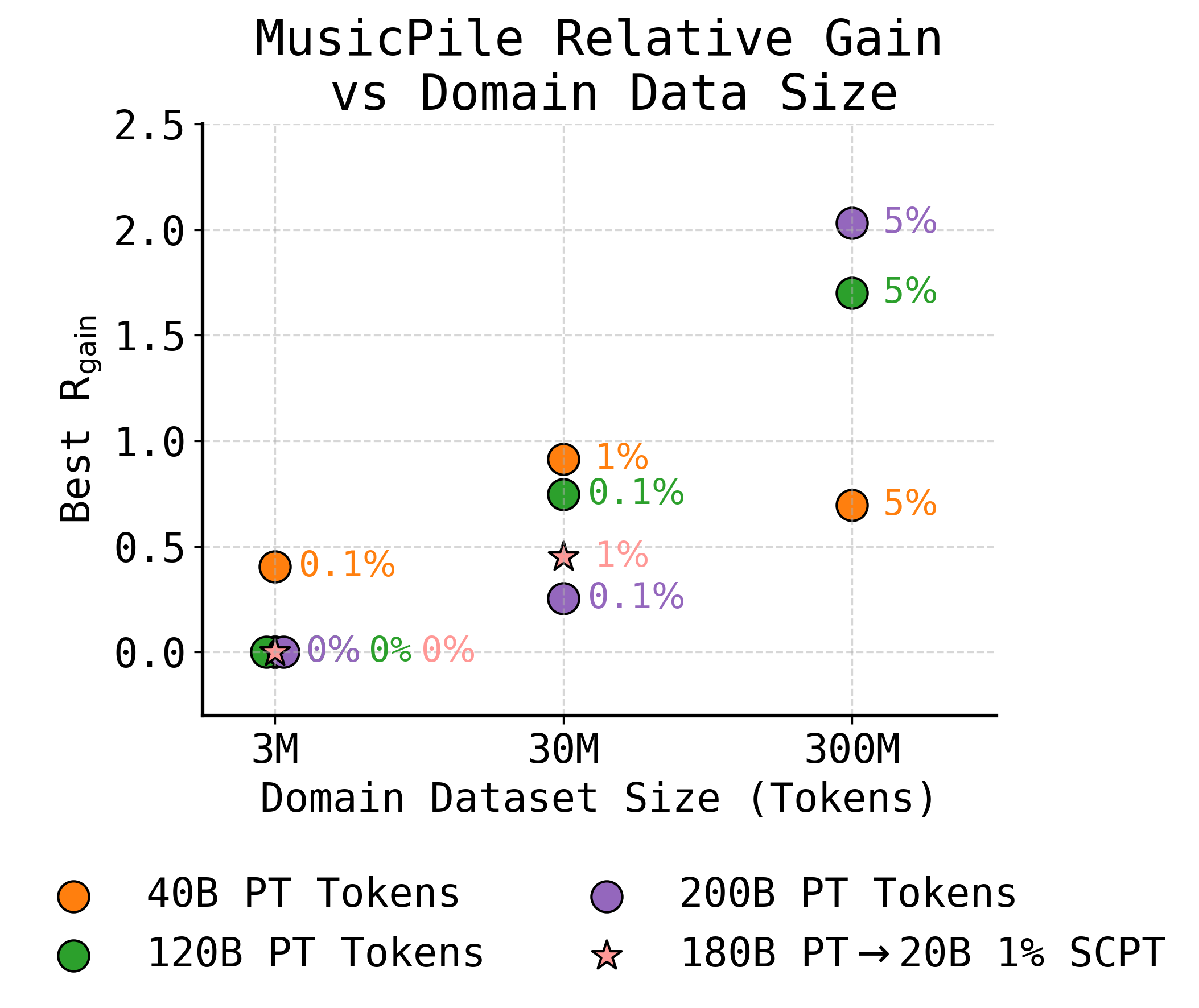}
    \end{minipage}
    \begin{minipage}[t]{0.44\linewidth}
    \vspace{0pt}
    \includegraphics[width=\linewidth]{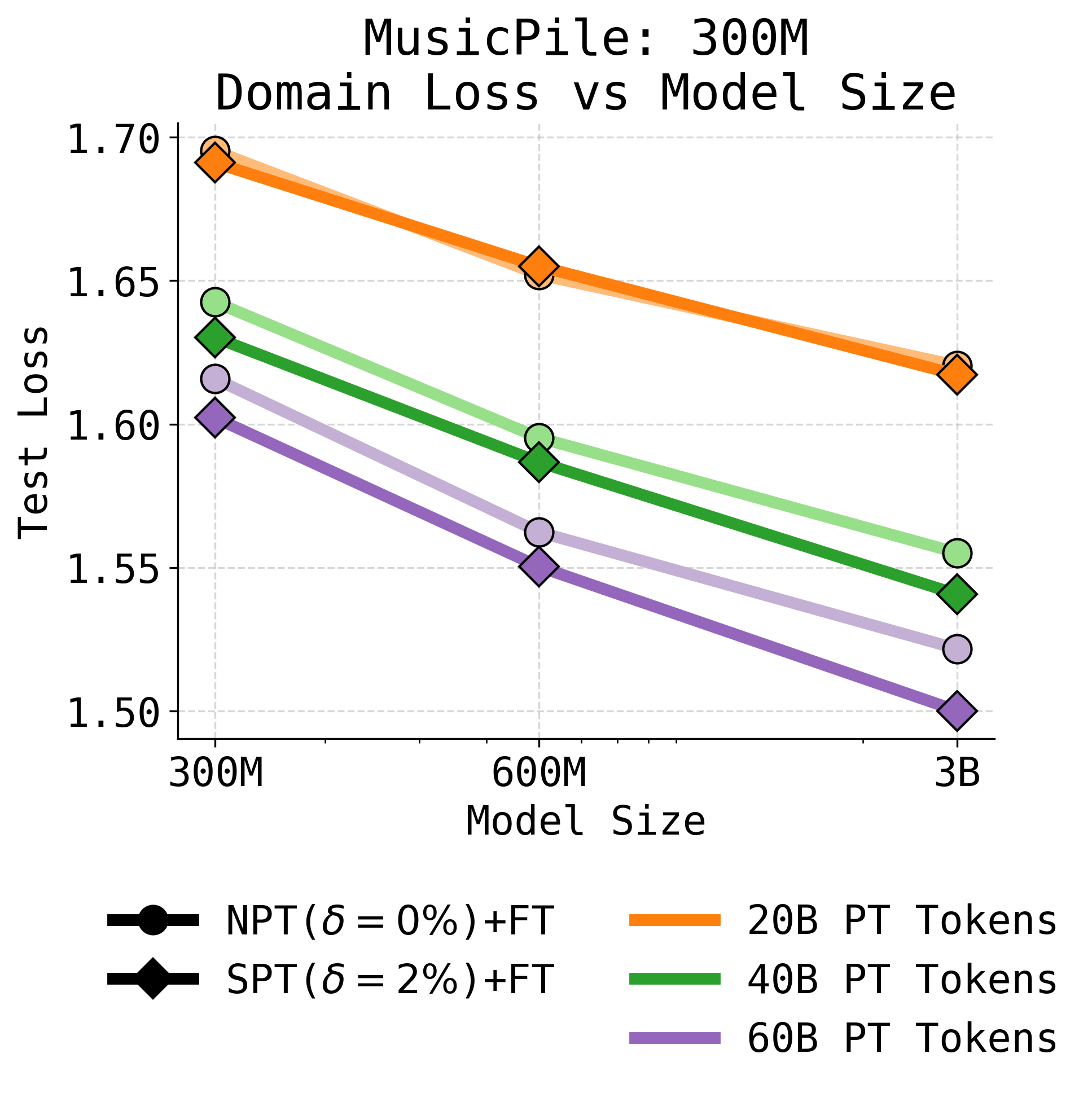}
    \end{minipage}
    \caption{\textbf{Effect of domain data size and model size on
    $\mathcal{R}_{\text{gain}}$.} \emph{Left:} Relative gain of \texttt{SPT} over
    \texttt{NPT} as a function of domain dataset size (\{3M, 30M, 300M\} tokens from
    MusicPile) at three pretraining scales (40B, 120B, 200B tokens). For the 300M-token
    dataset, gains are consistently high across all scales. For smaller datasets,
    \texttt{SPT} is less helpful at longer training horizons due to overfitting from
    excessive repetition; in these regimes, specialized continued pretraining (star)
    is preferable. \emph{Right:} Domain test loss vs.\ model size for \texttt{NPT}
    ($\delta=0\%$) and \texttt{SPT} ($\delta=2\%$) on MusicPile-300M. Relative gain grows with scale, driven not only by decreasing \texttt{NPT} loss but also a widening \texttt{SPT}--\texttt{NPT} gap.}
    \label{fig:size-vs-delta}
\end{figure}

\subsection{Domain Dataset Size}
\label{subsec:dataset-size}

Even when overlap between pretraining and target data is substantial, repeated exposure
to a limited domain corpus can induce overfitting. We examine how the magnitude of
\texttt{SPT}'s relative gain varies as a function of domain dataset size, evaluating
mixture fractions $\delta \in \{0, 0.1, 1, 2, 5\}$ across subsets of \{3M, 30M, 300M\}
tokens from MusicPile. Figure~\ref{fig:size-vs-delta} reports the maximum
$\mathcal{R}_{\text{gain}}$ achieved at multiple pretraining token budgets.

At a modest pretraining scale of 40B tokens, \texttt{SPT} yields consistent gains
($\mathcal{R}_{\text{gain}} \geq 0.5\%$) across all dataset sizes, with the 30M-token
subset exhibiting the largest improvement. However, as the pretraining budget increases
to 120B tokens and beyond, the regimes diverge. For the 300M-token dataset, relative
gain continues to increase with scale. In contrast, for smaller datasets (3M--30M
tokens), the benefit diminishes and can become negative. In these settings, even small
mixture fractions (e.g., $\delta=0.1\%$) induce excessive repetition, leading to
overfitting during pretraining and degraded post-finetuning performance.

In such extremely data-constrained regimes (on the order of tens of millions of tokens or
fewer), we find that \emph{specialized continued pretraining} (\texttt{SCPT}), in which
domain data is introduced at later stages of pretraining, may still be more effective than \texttt{NPT}.
Concretely, we take the 180B-token \texttt{NPT} checkpoint and continue training for
an additional 20B tokens with a $1\%$ domain mixture (pink stars in Figure \ref{fig:size-vs-delta}). For the 30M-token dataset,
\texttt{SCPT} achieves higher relative gain than full \texttt{SPT} from initialization. For the 3M-token dataset, \texttt{SCPT} at 1$\%$ still overfits excessively.

These results indicate that the interaction between repetition and dataset size induces
distinct scaling regimes: when the domain corpus is sufficiently large (e.g., 300M
tokens), early integration during pretraining yields sustained benefits; when the corpus
is small, deferring domain mixing to later stages improves generalization.

\subsection{Model Size}
The relative gains from \texttt{SPT} persist across model scales and actually increase
with parameter count. Figure~\ref{fig:size-vs-delta} (right) plots domain test loss as a
function of model size for \texttt{NPT} ($\delta=0\%$) and \texttt{SPT} ($\delta=2\%$)
across multiple pretraining budgets on MusicPile-300M. Across all settings, the gap in
test loss between \texttt{SPT} and \texttt{NPT} widens with model size: the 3B model
exhibits the largest reduction in post-finetuning test loss under \texttt{SPT}.

We believe this trend is consistent with the overfitting interpretation. Larger models
have greater representational capacity and are therefore more prone to memorizing a
limited domain corpus during finetuning. Collectively, these findings show that the
benefits of \texttt{SPT} amplify with model scale. As capacity increases, so does the
relative advantage of integrating domain data during pretraining, highlighting that
\texttt{SPT} becomes increasingly effective in the large-model regime.

\subsection{Pretraining compute budget}
\label{subsec:compute-budget}

\begin{figure}[t]
\centering
\includegraphics[width=0.8\linewidth]{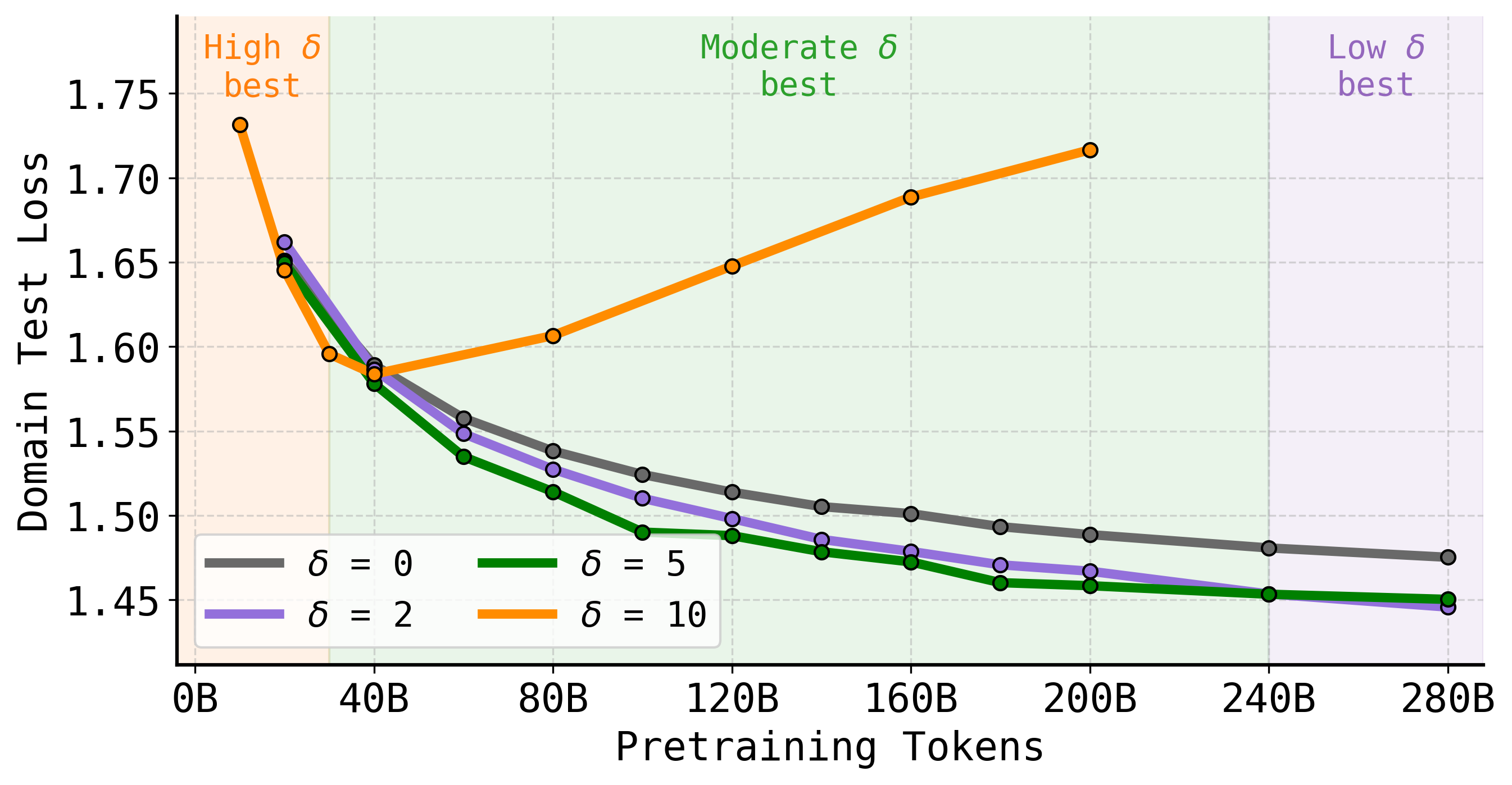}
\caption{\textbf{The optimal mixture shifts with compute budget.} Domain test loss on
MusicPile throughout \texttt{SPT} for mixture fractions $\delta \in \{0\%, 2\%, 5\%,
10\%\}$. At small compute budgets (left region), high domain mixtures ($\delta=10\%$)
achieve the lowest test loss. As training progresses, these mixtures overfit and their
loss increases, while moderate mixtures ($\delta=5\%$) become optimal (middle region).
Past 240B tokens, lower mixtures ($\delta=2\%$) are best. Regardless of budget,
\texttt{SPT} with an appropriate $\delta$ outperforms \texttt{NPT} ($\delta=0\%$).}
\label{fig:compute-budget}
\end{figure}

The optimal mixture fraction $\delta$ also depends on the available pretraining compute.
To study this interaction, we track domain test loss throughout training for mixture
fractions ranging from $0\%$ to $10\%$ on MusicPile (Figure~\ref{fig:compute-budget}).

At smaller pretraining budgets (e.g., under 30B tokens), larger mixture fractions perform
best, with $\delta=10\%$ achieving the lowest test loss. However, as training progresses,
these large mixtures begin to overfit due to repeated exposure to the domain corpus. By
approximately 50B tokens, the $\delta=10\%$ curve begins to degrade while moderate
mixtures (e.g., $\delta=5\%$) continue to improve. At longer training horizons (200B
tokens), the best performance is achieved by smaller mixtures in the range of
$\delta=2$--$5\%$. Crucially, regardless of the pretraining budget, \texttt{SPT} with an
appropriate choice of $\delta$ outperforms \texttt{NPT}.

These results demonstrate that the optimal degree of domain mixing is compute-dependent.
Larger mixture fractions are advantageous when training budgets are limited, as they
accelerate learning of domain structure. At larger compute scales, however, excessive
repetition leads to overfitting, shifting the optimal regime toward smaller mixture
fractions. Consequently, practitioners should treat $\delta$ as a function of training
horizon: higher mixtures for short pretraining runs and lower mixtures for longer ones.

\section{Does Specialized Pretraining Help Under Replay as Well?}
\label{sec:replay}
\begin{figure}[t]
\centering
\includegraphics[width=0.8\linewidth]{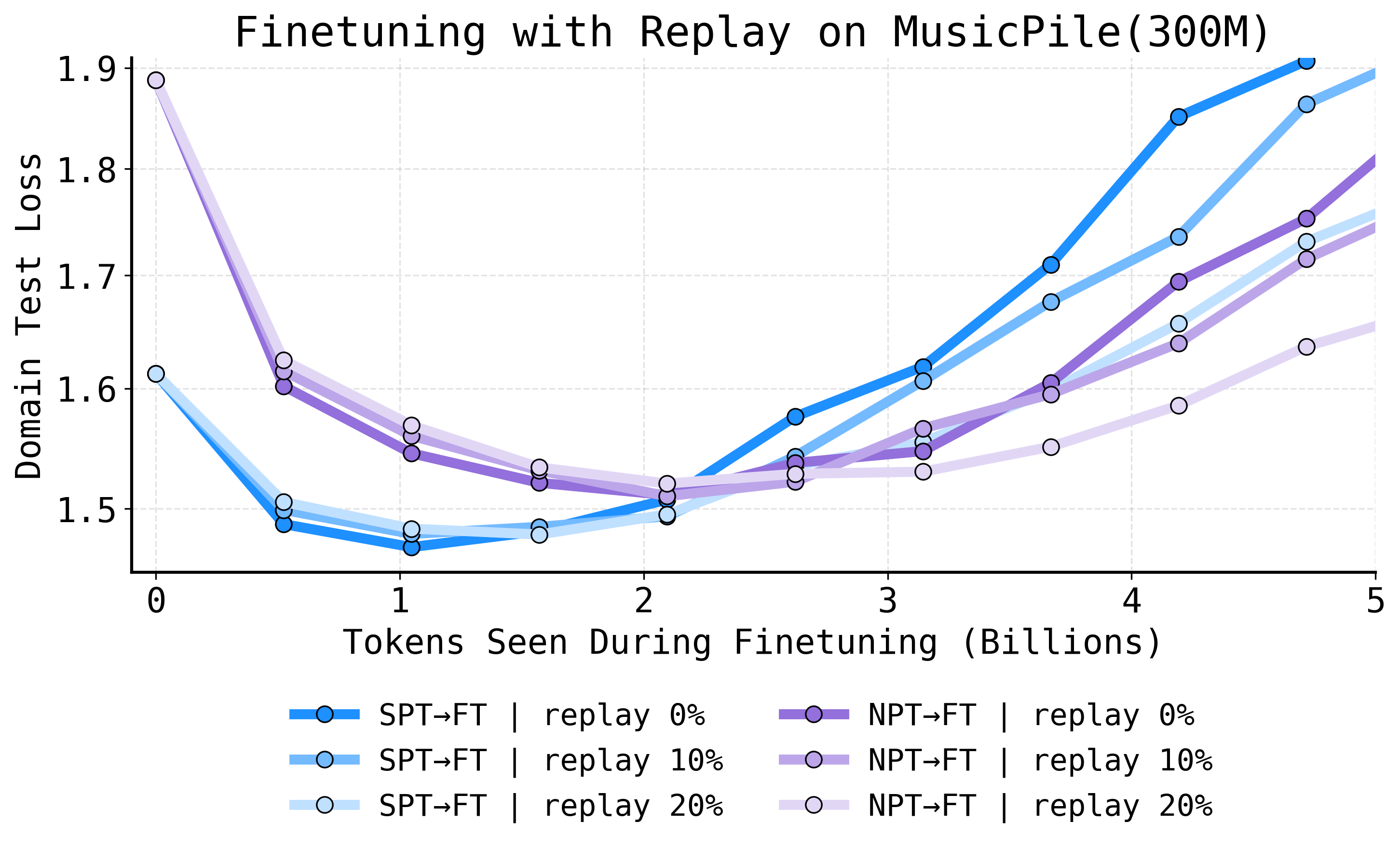}
\caption{\textbf{Replay is not a substitute for early domain exposure.}
MusicPile validation-loss trajectories during replay-based continued pretraining.
Blue lines (\texttt{SPT} $\to$ \texttt{CPT}) consistently achieve lower domain test loss than purple lines (\texttt{NPT} $\to$ \texttt{CPT}) across all replay rates (lighter shades indicate higher replay: $0\%\rightarrow10\%\rightarrow20\%$). While higher replay slows overfitting, it does not close the gap between \texttt{SPT} and \texttt{NPT}, confirming that \emph{when} domain data is seen has a lasting impact on the model's performance on specialized domains.
}
\label{fig:musicpile_replay_curves}
\end{figure}

Replay is commonly used during finetuning to mitigate forgetting by mixing previously seen data back into training \citep{parmar2024cpt, blakeney2024doesdatasparkjoy, kotha2026replay, liu2025midtraining}.
If replay already reintroduces general data during finetuning, an important question is whether replay has a similar regularizing effect or whether early exposure to domain data through \texttt{SPT} still outperforms \texttt{NPT}. 
We compare \texttt{SPT} $\to$ \texttt{FT} against \texttt{NPT} $\to$ \texttt{FT}, where replay-based FT mixes MusicPile (domain) with a Dolma replay mixture (general data).
We evaluate replay rates $\{0\%,10\%,20\%\}$ tuning the learning rate separately for each setting, and report MusicPile test loss across the finetuning trajectory (Figure~\ref{fig:musicpile_replay_curves}).

Across all replay settings, \texttt{SPT} $\to$ \texttt{FT} consistently achieves lower domain test loss than \texttt{NPT} $\to$ \texttt{FT}, reinforcing the core thesis that \emph{when} domain data is seen matters. Notably, 10\% replay helps \texttt{NPT}, but \texttt{NPT} $\to$ \texttt{FT} falls well short of \texttt{SPT} $\to$ \texttt{FT} with no replay. We hypothesize that these two forms of data mixing, diffuse domain exposure during pretraining versus general-data replay during finetuning, induce qualitatively different effects. As shown in Section~\ref{subsec:overfitting}, \texttt{SPT}'s benefit is more implicit, surfacing only after finetuning. The precise mechanism behind this asymmetry remains an open question.

\section{Predicting Overfitting with Scaling Laws}
\label{sec:scaling-laws}
Specialized pretrained models see multiple repetitions of finetuning data starting from pretraining. As a result, the optimal mixture percentage $\delta$ depends on the pretraining compute budget. While larger $\delta\%$ is optimal at shorter pretraining scales, there exists a threshold beyond which test loss plateaus or degrades due to excessive repetition. This observation motivates the following question:
\begin{center}\emph{Can we predict the domain test loss across specialized pretraining and finetuning as a function of $\delta$?}
\end{center}

Addressing this question is challenging since we explicitly have to model the overfitting regime, which the standard power law cannot anticipate. Furthermore, it's unclear 
how to predict scaling laws post-finetuning. We divide modeling overfitting scaling laws into two steps. First, we separately model how domain train and test losses scale during SPT as a function of $\delta$. Second, we measure the difference in test loss before and after finetuning, which we find also scales reliably with the pretraining tokens.

\begin{figure*}[t]
    \centering
        \centering
        \begin{minipage}[t]{0.63\textwidth}
        \vspace{1mm}
        \includegraphics[width=\textwidth]{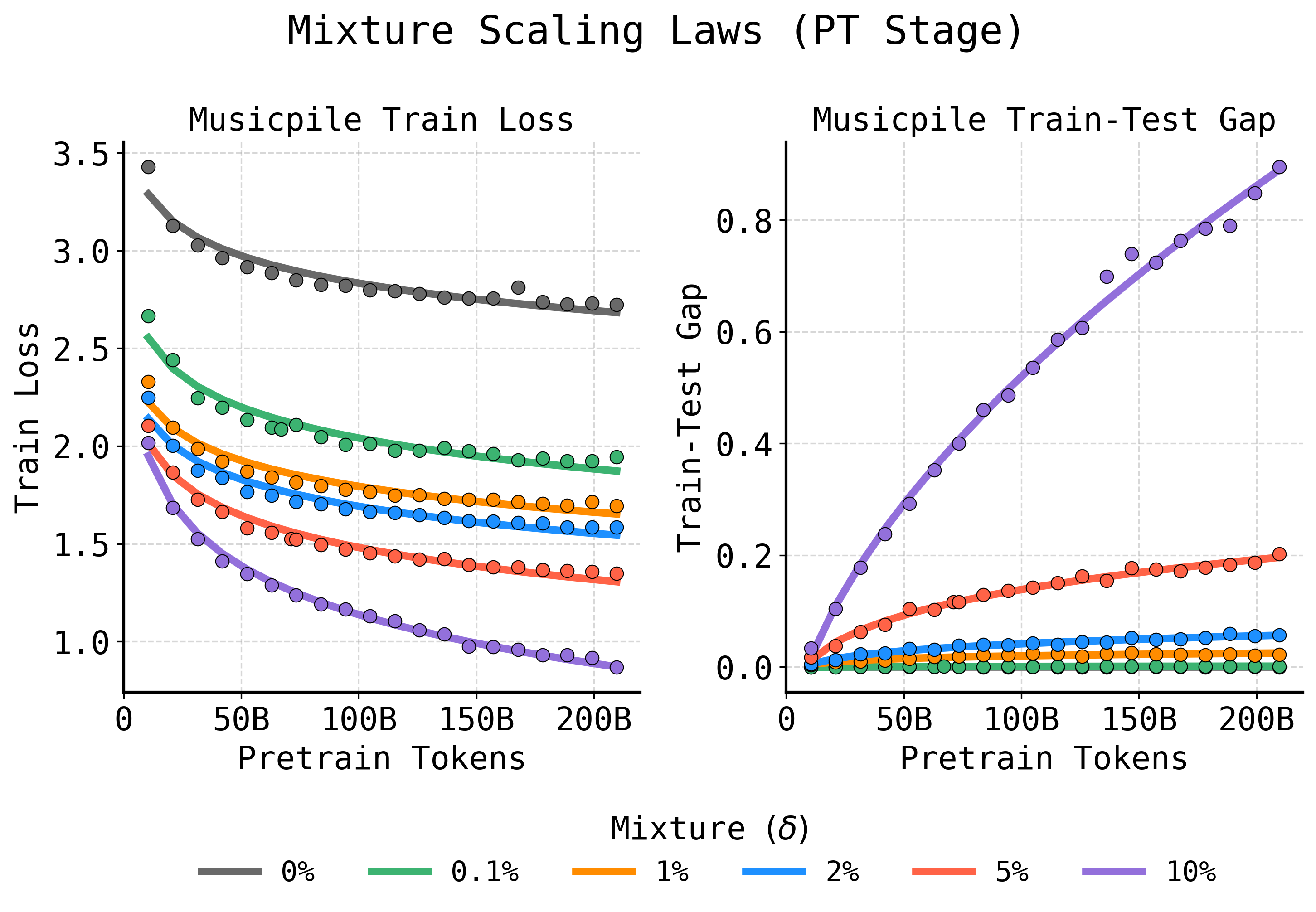}
        \end{minipage}
        \begin{minipage}[t]{0.36\textwidth}
        \vspace{1mm}
        \includegraphics[width=\textwidth]{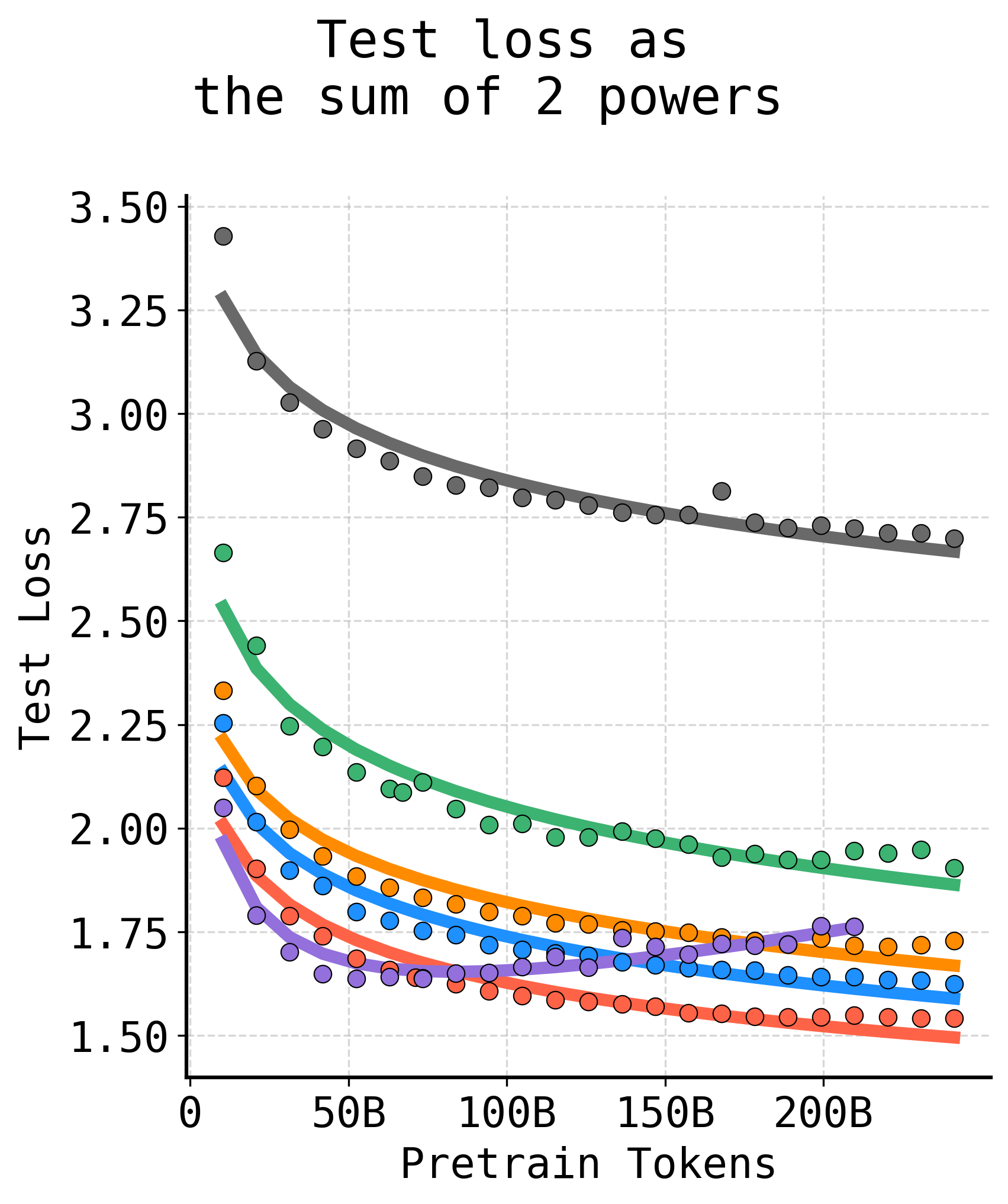} 
        \end{minipage}
    
    \caption{\textbf{Repetition scaling laws as the sum of two powers.} As the domain data is repeated throughout pretraining during specialized pretraining, for high mixture percentages, the model may overfit and the test loss goes back up. A single power law cannot model this overfitting stage. Instead, we fit a power law separately, for the domain training loss and the domain train-test gap. The latter also follows a power law with a positive exponent. Then the test loss can be modeled using the sum of two powers. Furthermore, we model all coefficients as a function of the mixture percentage, according to Equation \ref{eq:scaling_law_constraints}. \label{fig:scaling_schematic}}
\end{figure*}
\subsection{Overfitting during Specialized Pretraining}
When repeating domain-specific data, the test loss improves up to a certain number of repeats, after which the model overfits. Previous works that study scaling laws for repeated data lack an expression that accurately models the overfitting stage \citep{kaplan2020scaling_clean, Goyal_2024_CVPR, muennighoff2025scalingdataconstrainedlanguagemodels}. These works often express overfitting as a power law whose exponent is itself modeled by a decaying function of the number of repetitions, introducing nested nonlinearities that make the resulting expression difficult to fit from limited data and hard to extrapolate beyond the observed training range.

Instead, we find that a simple alternative model does the trick. Instead of fitting a single power law, we decompose the test loss into the training loss and the train-test gap and model them as power laws separately. While the training loss can be modeled using the usual power law with a negative exponent, the train-test gap monotonically increases with more repeats and can be modeled using a positive exponent. Then the test loss is simply the sum of the two terms. We illustrate this across all our SPT runs over MusicPile in Figure~\ref{fig:scaling_schematic}.

Furthermore, we model each learned coefficient as a function of the mixture fraction $\delta$. In total, we propose a scaling law for the \emph{domain} test loss with respect to the number of pretraining tokens $T$:
\begin{align}
\label{eq:scaling_laws}
\mathcal{L}_{\mathrm{train}}(T,\delta)
&= A_{\mathrm{train}}
   \,T^{\,b_{\mathrm{train}}(\delta)}
 + C_{\mathrm{train}}(\delta),\\
\mathcal{L}_{\mathrm{gap}}(T,\delta)
&= A_{\mathrm{gap}}(\delta)
   \,T^{\,b_{\mathrm{gap}}(\delta)},\\
\mathcal{L}_{\mathrm{test}}(T,\delta)
&= \mathcal{L}_{\mathrm{train}}(T,\delta)
 + \mathcal{L}_{\mathrm{gap}}(T,\delta),
\end{align}
where
\begin{align}
\label{eq:scaling_law_constraints}
b_x(\delta) &= \delta b_{x,s}+(1-\delta)b_{x,g}, \qquad x\in\{\mathrm{train},\mathrm{gap}\}\\
\label{eq:gamma} A_{\mathrm{gap}}(\delta) &= \alpha_1 \delta^{\alpha_2}\exp(\alpha_3\delta),\\
\label{eq:loglinear} C_{\mathrm{train}}(\delta) &= \kappa_0-\kappa_1\log(\delta+\kappa_2)-\kappa_3\delta.
\end{align}

For the training loss, we fix $A_{\mathrm{train}}$ to be a fixed constant, while we model it the Gamma kernel function for the gap. Furthermore, we use a log linear expression to model $C_{\mathrm{train}}(\delta)$, which we find best fits the experimental data. We provide further evidence for these design choices in Appendix \ref{app:overfitting}. 

\paragraph{Interpreting the exponent} The exponent in particular has an interpretable form. The rate at which the specialized domain training loss decreases can be modeled by both the general pretraining data and the specialized domain training data having (negative) utility $b_g$. For the training loss, both exponents are negative $b_g, b_s < 0$ while $|b_g| < |b_s|$, to represent that the domain training loss goes down slower for smaller mixture fractions. On the other hand, for the train-test gap, we tradeoff between $b_g < 0$ and $b_s > 0$ to represent how higher mixture fractions accelerate overfitting. 

The key insight is that $b_{\mathrm{train}}(\delta)$ is strictly negative, indicating that training loss decreases monotonically throughout pretraining. Notably, the train-test gap also follows a power law, but with a positive $b_{gap}$, meaning that overfitting increases monotonically with more epochs over the dataset. However,$b_{gap}$  remains below 1 across all mixture percentages up to 10\%, implying that overfitting grows sublinearly over extended pretraining horizons. This accounts for the sustained effectiveness of SPT over prolonged training.
\begin{table}
\end{table}

\begin{figure}[t]
    \centering
    \begin{minipage}[b]{0.39\textwidth}
    \vspace{1em}
    \includegraphics[width=\textwidth]{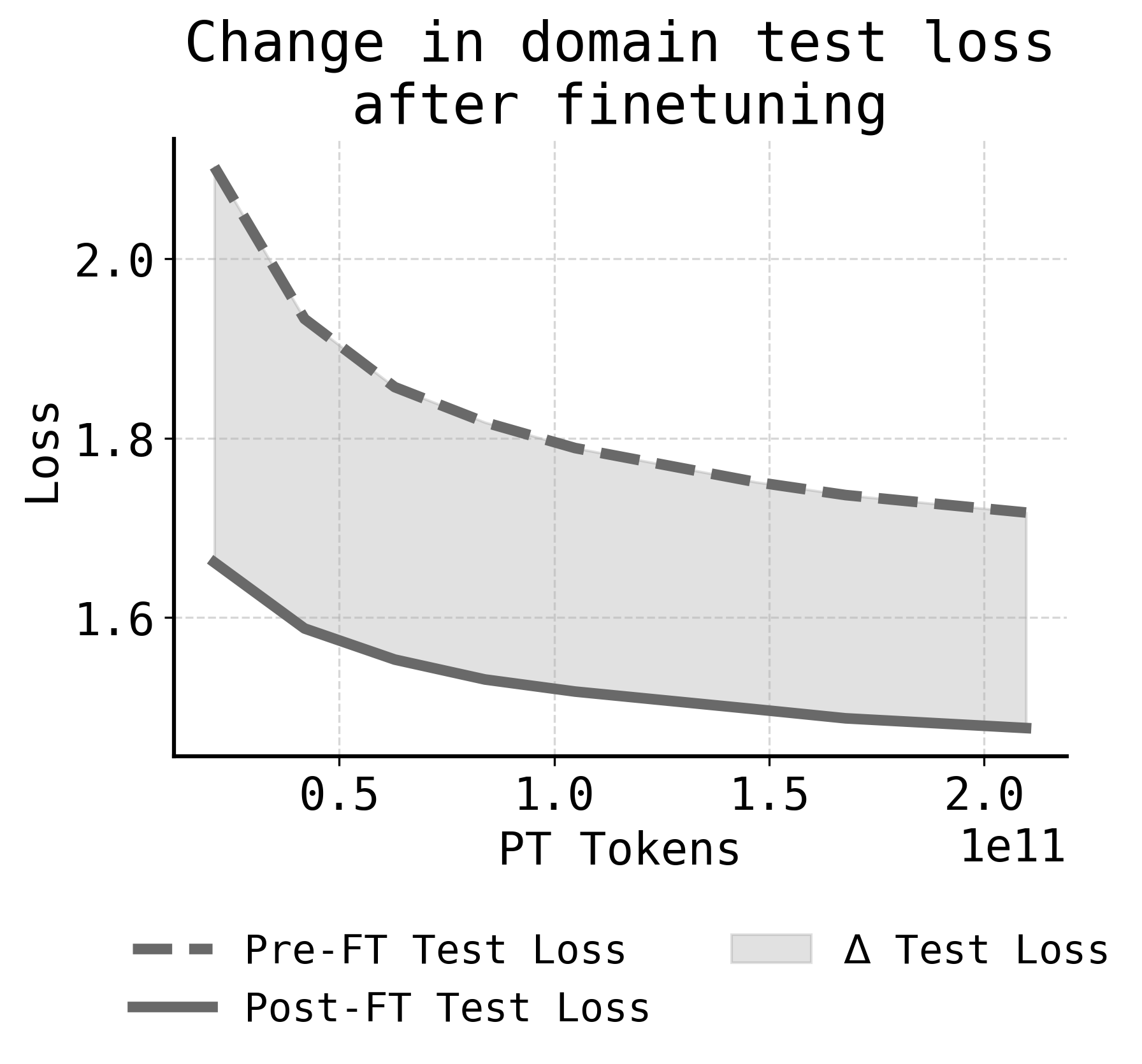}
    \end{minipage}
    \begin{minipage}[b]{0.6\textwidth}
    \vspace{-1mm}
    \includegraphics[width=\textwidth]{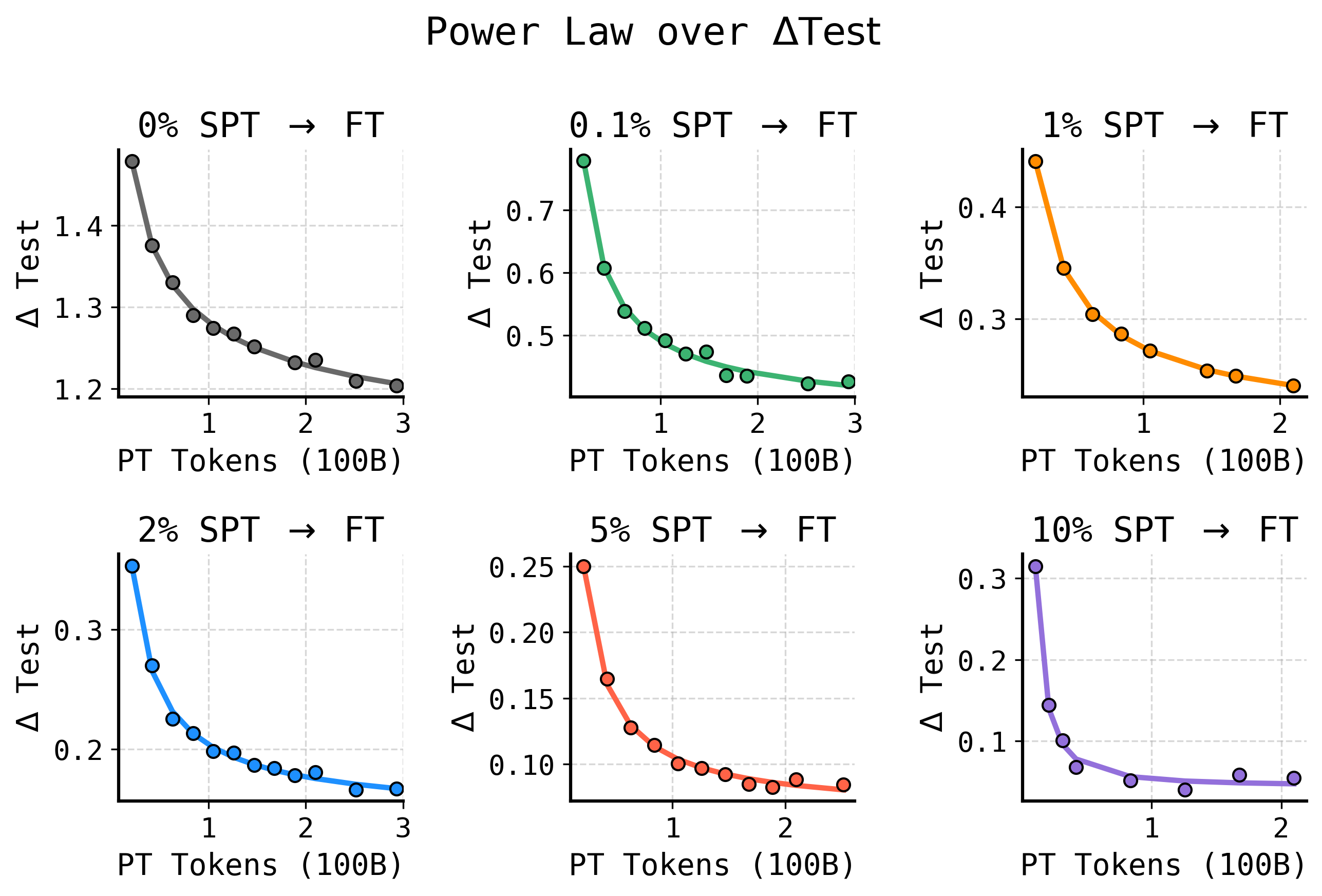}
    \end{minipage}
    \caption{\textbf{Predicting loss after finetuning.} We measure the difference in domain-specific test loss right after pretraining versus the best test loss after finetuning ($\Delta \ \mathrm{Test}$) as a function of the pretraining tokens. We observe that this difference follows a power law relationship. \label{fig:delta_test}}
\end{figure}
\subsection{Predicting Test Loss After Finetuning}

Now that we have established an accurate model characterizing the domain-specific test loss as a function of pretraining compute, we turn our attention to predicting the test loss after finetuning. Directly modeling the post-finetuning test loss proves challenging, as it entangles the contributions of pretraining and finetuning. Instead, we leverage our existing pretraining loss model and separately characterize the marginal effect of finetuning by modeling the difference between the test loss at the pretrained checkpoint and the test loss after subsequent finetuning.

Formally, we define the change in domain test loss after finetuning as \begin{align} 
\Delta \ell_{test} = \ell_{test}({\theta_{PT}}) - \ell_{test}({\theta_{PT+FT}}),
\end{align} 
where $\ell_{test}({\theta_{PT}})$ denotes the test loss of the pretrained model and $\ell_{test}({\theta_{PT+FT}})$ denotes the best loss after the model has been further finetuned on the target domain. A positive value of $\Delta \ell_{test}$ thus indicates that finetuning has reduced the test loss relative to the pretrained checkpoint. Since we employ early stopping, the difference will always be above 0.

Empirically, we observe that this change in test loss follows a remarkably consistent power-law relationship as a function of the number of pretraining steps
T: $\Delta \ell_{test} = a T^b + c$. Intuitively, this relationship captures how the benefit conferred by finetuning varies depending on the stage of pretraining at which the model is finetuned. We illustrate the power-law relationship in MusicPile in Figure \ref{fig:delta_test} and provide supporting evidence for the other domains in Appendix \ref{app:overfitting}.

\begin{figure}[t]
\begin{subfigure}[b]{0.49\textwidth}
\vspace{0.1em}
\includegraphics[width=\textwidth]{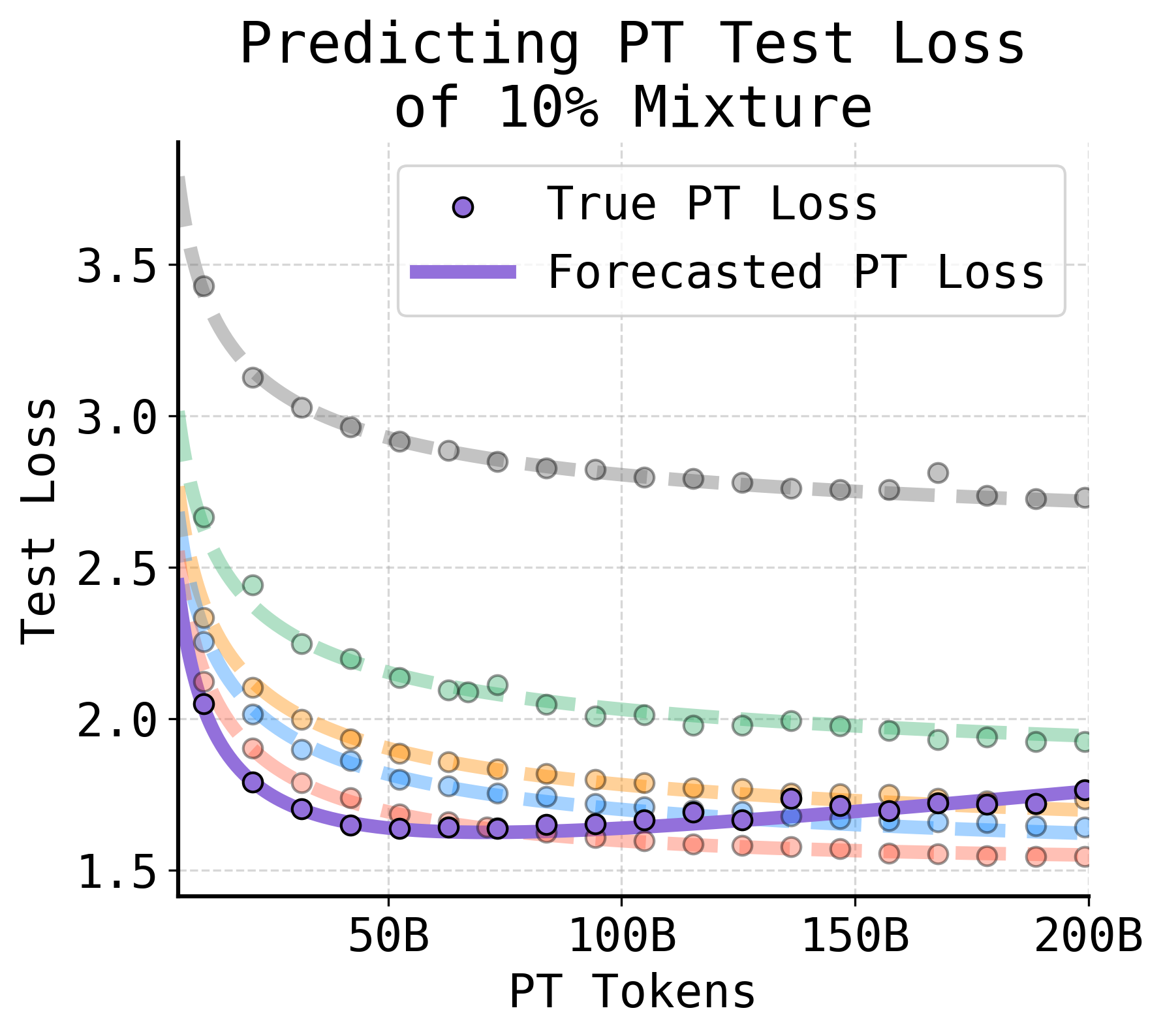}
\caption{Predicting $10\%$ Pretraining Curve \label{fig:forecast_pt}}
\end{subfigure}
\begin{subfigure}[b]{0.48\textwidth}
\vspace{0.1em}
\includegraphics[width=\textwidth]{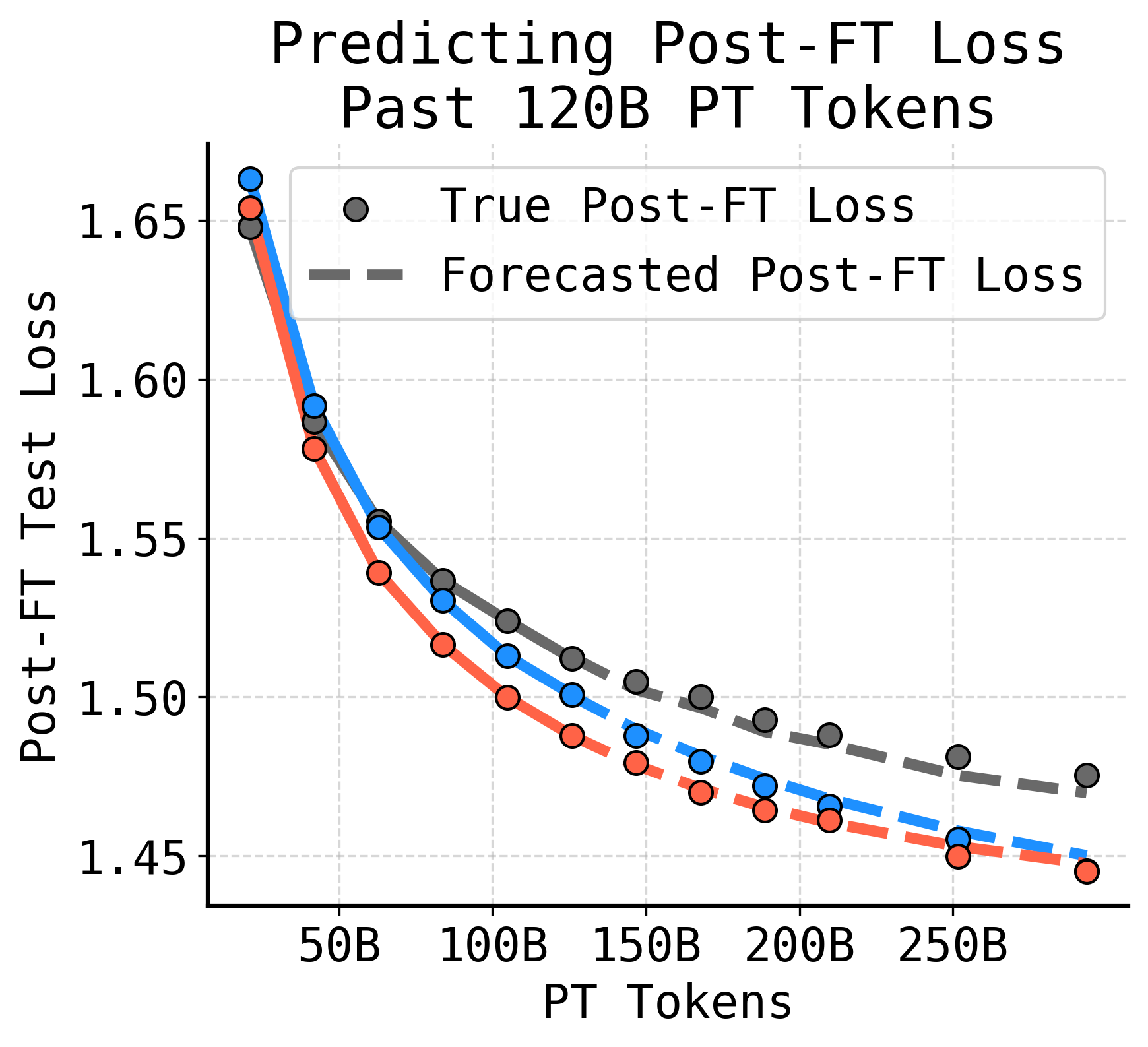}
\caption{Post-FT Loss past 120B PT Tokens \label{fig:forecast_ft}}
\end{subfigure}
\caption{\textbf{Forecasting domain loss and optimal mixture from small-scale runs.} We demonstrate two ways our scaling laws can guide the choice of mixture percentage for SPT. (Left) We predict the entire domain test loss trajectory of $10\%$ SPT by fitting our scaling laws (Equation \ref{eq:scaling_laws}) using only the curves from $\delta \in [0, 5]$. The forecast correctly predicts that $10\%$ SPT begins to overfit around 70B tokens. (Right) We extrapolate post-finetuning test loss past 120B pretraining tokens and correctly predict that $2\%$ SPT surpasses $5\%$ SPT around 280B tokens. Together, these forecasts allow practitioners to identify the right mixture fraction without running the full training sweep.}
\end{figure}
\subsection{Forecasting Examples}
Together, our scaling laws precisely model the domain test loss when repeating a small set of data multiple times across pretraining \emph{and} finetuning. We demonstrate two ways our scaling laws can be used to guide the choice of optimal mixture percentage for SPT. 
Recall that we model the domain test loss during SPT with any mixture percentage as the sum of two powers with the exponent for both powers set to $(1-\delta) b_g + \delta b_s$ where $b_g, b_s$ are learned variables. We learn $b_g$ and $b_s$ and the other variables for the other terms $A(\delta)$ and $C(\delta)$ using the mixtures $\delta \in [0, 5]$, then extrapolate the full domain loss curve of SPT for 200B tokens. In Figure~\ref{fig:forecast_pt}, we compare our prediction to the true $10\%$ SPT loss curve. The forecast tracks the observed values closely across the full 200B-token range, correctly predicting that SPT at $10\%$ begins to overfit around 70B tokens.

Moreover, we can combine our pretraining scaling laws with our power law relationship over $\Delta \ell_{\mathrm{test}}
$ to fully extrapolate post-finetuning loss beyond 120B pretraining tokens for $\delta \in [0, 5\%]$. As shown in Figure~\ref{fig:forecast_ft}, our extrapolation correctly predicts that $2\%$ SPT surpasses $5\%$ SPT around 280B tokens (consistent with the full runs in Figure~\ref{fig:compute-budget}). In practice, these scaling laws eliminate the need for expensive sweeps over mixture fractions and training horizons. A small number of pilot runs suffices to fit the expressions and select the right SPT configuration for a given compute budget, making specialized pretraining both effective and predictable.

\section{Discussion}

Our work exposes \emph{the finetuner's fallacy}. The standard approach to domain
adaptation, finetuning a large general-purpose model appears cheap because it avoids
the cost of pretraining. But this accounting ignores inference. A 1B \texttt{SPT} model
costs more to train than finetuning a 3B model, but the $3\times$ smaller model is
cheaper to serve (Figure~\ref{fig:finetuners_tax}). At scale, the break-even point
arrives quickly, after which \texttt{SPT} saves both compute and money while delivering
superior performance. For deployed models, inference dominates total energy consumption,
so enabling the same capability with fewer parameters also reduces the carbon footprint
of domain-adapted language models.

Importantly, the finetuner's fallacy is not ``fine-tuning is always enough'' or
``always pretrain your model.'' Real systems occupy different points on a
continuum, and the right strategy depends on how much domain data you have, how far your
domain sits from web text, and what your serving constraints look like. Our experiments
provide systematic evidence for where the transitions between these regimes lie.
SPT's gains are largest when the target domain is far from web text, scale with model size, and
depend on the interaction between domain dataset size and pretraining budget. Our overfitting scaling
laws let practitioners navigate this tradeoff from a small number of pilot runs rather
than exhaustive search.

The broader lesson is simple: scarce and specialized domain data should not be treated as a final-stage
resource. Across three domains ranging from symbolic music to mathematical proofs,
mixing even 1--5\% domain data into pretraining consistently outperformed the standard
pipeline of general pretraining followed by finetuning. These gains persisted even under
replay-based continued pretraining, confirming that \emph{when} domain data enters the
training pipeline has a lasting impact on model performance. In 2026, the cost of pretraining
continues to fall, and the case for early integration only strengthens with it. As
organizations increasingly seek to deploy models tailored to proprietary domains, these
findings point toward a shift in how specialized models should be trained: away from
post-training patches applied to generic checkpoints, and toward natively specialized
models that incorporate domain knowledge from the start.

\section*{Contributions}
\label{sec:contri}

Christina Baek led the project and conducted all the core experiments. 
Pratyush Maini provided project direction and contributed to the experimental design and analysis.

David Schwab, Ricardo Monti, Aditi Raghunathan, Zico Kolter, Bogdan Gaza, Ari Morcos, and Matthew Leavitt 
provided guidance throughout the project and feedback on the draft.

The Datology team contributed to helpful discussions and provided the infrastructure that supported the experiments in this paper: 
Amro Abbas, Rishabh Adiga, Cody Blakeney, Maximilian Böther, Paul Burstein, Aldo Gael Carranza, Alvin Deng, Parth Doshi, Vineeth Dorna, Alex Fang, Tony Jiang, Siddharth Joshi, Brett W. Larsen, Jason Chan Lee, Katherine L. Mentzer, Luke Merrick, Haakon Mongstad, Fan Pan, Anshuman Suri, Darren Teh, Jason Telanoff, Jack Urbanek, Zhengping Wang, Josh Wills, and Haoli Yin.

We thank Liz Gatapia for her help with the logo design.

\section*{Acknowledgements}
We'd like to thank Suhas Kotha, Jacob Springer, Gaurav Ghosal, and Lawrence Feng for their insights on finetuning and their feedback on earlier versions of this work.

\bibliographystyle{abbrvnat}
\bibliography{main}

\newpage
\appendix
\onecolumn
\section{Related Works}
\subsection{Catastrophic Forgetting and Replay in Domain Adaptation}

Catastrophic forgetting, the phenomenon whereby learning new information overwrites 
previously acquired knowledge, has been studied since at least \citet{mccloskey1989catastrophic}.
In the LLM era, \citet{gururangan2020dont} showed that continued pretraining on domain-relevant text improves downstream performance but can degrade general capabilities.
\citet{luo2023empirical} demonstrated that forgetting intensifies as model scale increases from 1B to 7B parameters during continual finetuning, and that general instruction tuning prior to specialization can mitigate the problem.
\citet{gupta2023continual} studied how to ``rewarm'' learning rate schedules for continual pretraining without destabilizing the model, a practical consideration our work shares.
\citet{shi2024continual} provide a comprehensive survey of continual learning strategies for LLMs, cataloguing replay-based, regularization-based, and architecture-based mitigation approaches.

Replay, mixing previously seen general data back into later training stages, is the most widely adopted defense.
\citet{parmar2024cpt} propose a recipe for continued pretraining that mixes general and domain data to mitigate forgetting, and \citet{blakeney2024doesdatasparkjoy} show that domain upsampling at the end of training can yield performance gains.
More recently, \citet{kotha2026replay} show that replaying generic pretraining data during finetuning not only prevents forgetting of general knowledge but can actually \emph{improve} target-task performance, increasing data efficiency by up to $2.06\times$ for mid-training.
They further find that replay helps more when there is less target data present in the pretraining mix, directly complementing our finding that SPT's advantage is largest when the domain is underrepresented.
On the measurement side, \citet{harmon2025postforgetting} propose sample-wise metrics revealing that large per-example forgetting can hide beneath stable aggregate accuracy, and \citet{thede2026captrack} introduce CapTrack, a capability-centric framework showing that post-training forgetting extends well beyond factual knowledge loss to encompass drift in multilingual robustness, instruction following, and calibration.

Our work~(\S~\ref{sec:replay}) shows that replay during continued pretraining is not a substitute for early domain exposure: SPT's gains persist across all replay settings.
This is consistent with the broader message of these works that \emph{when} data appears in the training pipeline matters as much as \emph{whether} it appears.

\subsection{Scaling Laws for Data Mixing and Repeated Data}

A separate line of work develops predictive models for how data composition affects final loss.
\citet{kaplan2020scaling_clean} established the first scaling laws relating model size, dataset size, and compute to language modeling loss.
\citet{muennighoff2025scalingdataconstrainedlanguagemodels} extended these laws to data-constrained regimes where tokens must be repeated, finding that multiple epochs are beneficial up to a point but eventually yield diminishing returns; however, their expressions do not model the overfitting stage that we observe.
\citet{ye2024datamixing,Goyal_2024_CVPR} propose \emph{data mixing laws} that predict loss as a function of domain mixture proportions, enabling practitioners to optimize data blends from small pilot runs.
\citet{que2024dcpt} derive a domain-specific continual pretraining scaling law that predicts the optimal mixture ratio between general and domain corpora as a function of model size and data budget.

Our overfitting scaling laws (\S\ref{sec:scaling-laws}) complement these efforts by separately modeling the training loss and train-test gap as power laws with opposing exponents.
This decomposition captures the non-monotonic test loss trajectory that arises from heavy repetition of a small domain corpus, a regime not addressed by prior scaling law formulations.
We further parametrize all coefficients as functions of the mixture fraction $\delta$, enabling extrapolation across both mixture percentages and training horizons from a small set of pilot runs.

\subsection{The synergy between pretraining and post-training}

More broadly, a capability-oriented view of the pretraining-vs-post-training boundary is
emerging across several subfields. In safety alignment, \citet{maini2025safety} and
\citet{sam2025whensafety} show that behaviors acquired during pretraining are harder to remove
via post-training than the reverse, implying that the stage at which data is introduced
determines how durably it shapes the model. In multilingual settings, \citet{longpre2025atlasadaptivetransferscaling}
find that cross-language transfer during pretraining improves low-resource language
performance more effectively than later-stage exposure. And in the post-training literature,
\citet{chu2025sftrl} demonstrate that RL with outcome-based rewards generalizes to unseen
task variants whereas SFT memorizes training examples, a finding formalized by
\citet{shenfeld2025rlrazor} as \emph{RL's Razor}: on-policy RL is implicitly biased toward
the policy closest in KL divergence to the base model, which \citet{lai2025rft} confirm
mitigates forgetting across sequences of continual post-training tasks. The common thread
across all three settings is that the \emph{manner} in which data is introduced, not just its
quantity, determines whether the model memorizes or generalizes. SPT operationalizes this
principle at the pretraining stage: by interleaving domain tokens among general data, no
single batch is dominated by the scarce domain corpus, producing a regularization effect
analogous to RL's on-policy sampling.

\subsection{Specialized Pretraining in Deployed Systems}

Industry practice illustrates that the choice between pretraining from scratch and finetuning an existing model lies on a continuum indexed by data scale, domain shift, and deployment constraints \citep{labelyourdata2025pretrainFinetune,raga2024pretraining,aws2024continual,databricks2025customllm}.
Bloomberg trained BloombergGPT, a 50B-parameter model on roughly 700B tokens combining a 363B-token proprietary finance corpus with general data, arguing that a finance-aware model outperforms a stack of task-specific finetuned models \citep{bloomberg2023press,wu2023bloomberggpt}.
Character.AI operates both large custom conversational models trained from scratch \citep{nix2023characteraifromscratch,character2024inference} and aggressively finetuned open-source models using SFT, DPO, and RL \citep{charaiblog2025opensource}.
Cursor's Composer illustrates the opposite direction: RL-based post-training atop an existing coding backbone rather than training from scratch \citep{codecademy2025cursor2,belitsoft2025composer}.
As Cursor pursues increasingly agentic capabilities, it may encounter \textit{the finetuner's fallacy} anew, as finetuning a base model that lacks agentic traces will eventually yield diminishing returns.

These cases suggest \textit{the finetuner's fallacy} is not simply ``fine-tuning is always enough'' or ``everyone should train their own model.'' Real systems occupy different points on a continuum indexed by (i) how much \emph{domain-specific data} they control, (ii) how far their domain is from web-scale distributions, and (iii) \emph{scale and latency} requirements. Yet there is little public evidence on where the transitions between these regimes actually lie. The goal of this paper is to move from anecdotes to a systematic characterization of when additional pretraining is warranted, and when fine-tuning suffices, as a function of data size, domain similarity, and model scale.

\newpage
\section{Optimization Details}
\label{app:hyperparameters}
\subsection{Pretraining Hyperparameter Configurations}
We match OLMo-1B pretraining settings to the publicly documented configuration \citep{Groeneveld2023OLMo} where possible, including the optimizer, cosine learning rate schedule, and batch size. 

\begin{description}[leftmargin=0pt,labelindent=0pt]
\item[Architecture] OLMo-1B 
\item[Batch Size] 2048 
\item[Context Length] 2048 
\item[Weight Tying] True
\item[Gradient Clipping] 1.0
\item[Weight Decay] 0.1
\item[AdamW Betas] 0.9, 0.95
\item[AdamW Epsilon] 1e-5
\item[Learning Rate Scheduler] Cosine decay schedule from $4e\text{-}4$ to $4e\text{-}5$ over $2T$ tokens. We use the same scheduler but cutting off the pretraining stage at $200B$ tokens. 
\end{description}

\subsection{Finetuning Hyperparameter Configurations} 
During finetuning, we ablate learning rate and warmup steps, while keeping other hyperparameters fixed. We set weight decay to be fairly small and remove weight tying. 

\begin{description}[leftmargin=0pt,labelindent=0pt]
\item[Batch Size] 512
\item[Learning Rate] \{1e-5, 4e-5, 1e-4\} 
\item[Warmup] \{50, 100, 200\}
\item[Weight Tying] False
\item[Weight Decay] 1e-7 
\item[AdamW Betas] 0.9, 0.95
\item[AdamW Epsilon] 1e-5
\item[Learning Rate Scheduler] Constant with warmup.
\end{description}

\section{Construction of Domain Datasets}
\paragraph{MusicPile} We take MusicPile \citep{musicpile} and subsample from music-specific sources: 
\texttt{sander-wood/irishman}, \texttt{Generated with GPT-4}, and \texttt{constructed from OpenChat, IrishMAN and KernScores}. A large proportion of the final dataset test for musical composition in ABC notation, in addition to general music knowledge.

\paragraph{ChemPile} We take ChemPile \citep{chempile} and subsample from the reasoning, instruction, and education domains. 

\paragraph{ProofPile} We subsample the data from entire ProofPile \citep{proofpile} set.

\newpage
\section{Overfitting Scaling Laws Extended}
\label{app:overfitting}

\subsection{Justification for $A_{\mathrm{gap}}$ and $C_{\mathrm{train}}$} 
In Section \ref{sec:scaling-laws}, we modeled scaling laws as a function of the mixture fraction $\delta$. In this section, we validate our choices for $A_{\mathrm{gap}}$ and $C_{\mathrm{train}}$. $A_{\mathrm{gap}}$ was modeled using the Gamma kernel (Equation \ref{eq:gamma}), and $C_{\mathrm{train}}$ was modeled using the log-linear function. 

\begin{figure}
    \centering
    \includegraphics[width=0.9\textwidth]{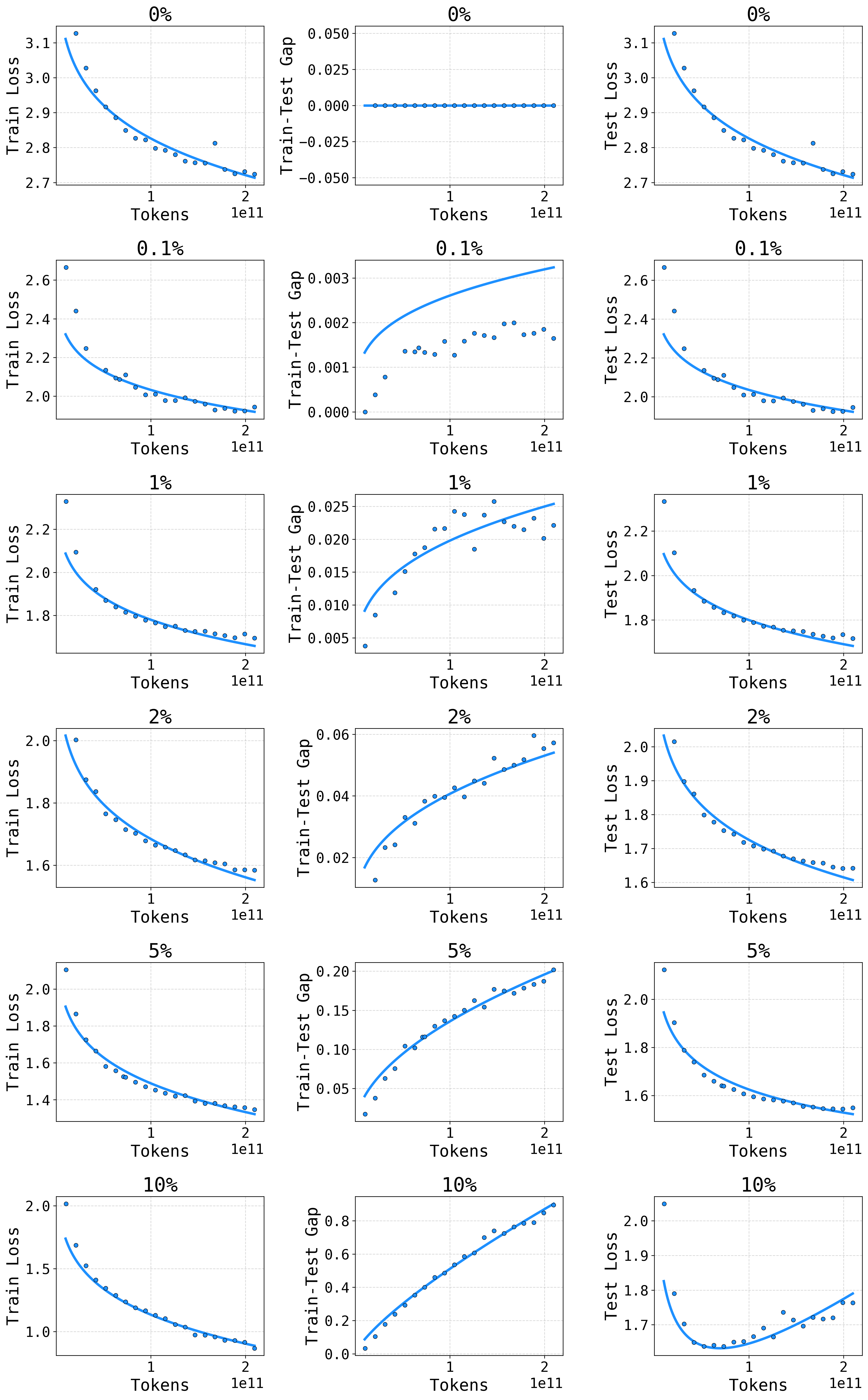}
    \caption{All power law fits over the training loss, train-test gap on \texttt{MusicPile-300M} over the course of SPT. Finally, test loss as the sum of two powers. \label{fig:musicpile_separate_fits}}
\end{figure}

\begin{figure}
    \centering
    \includegraphics[width=0.9\textwidth]{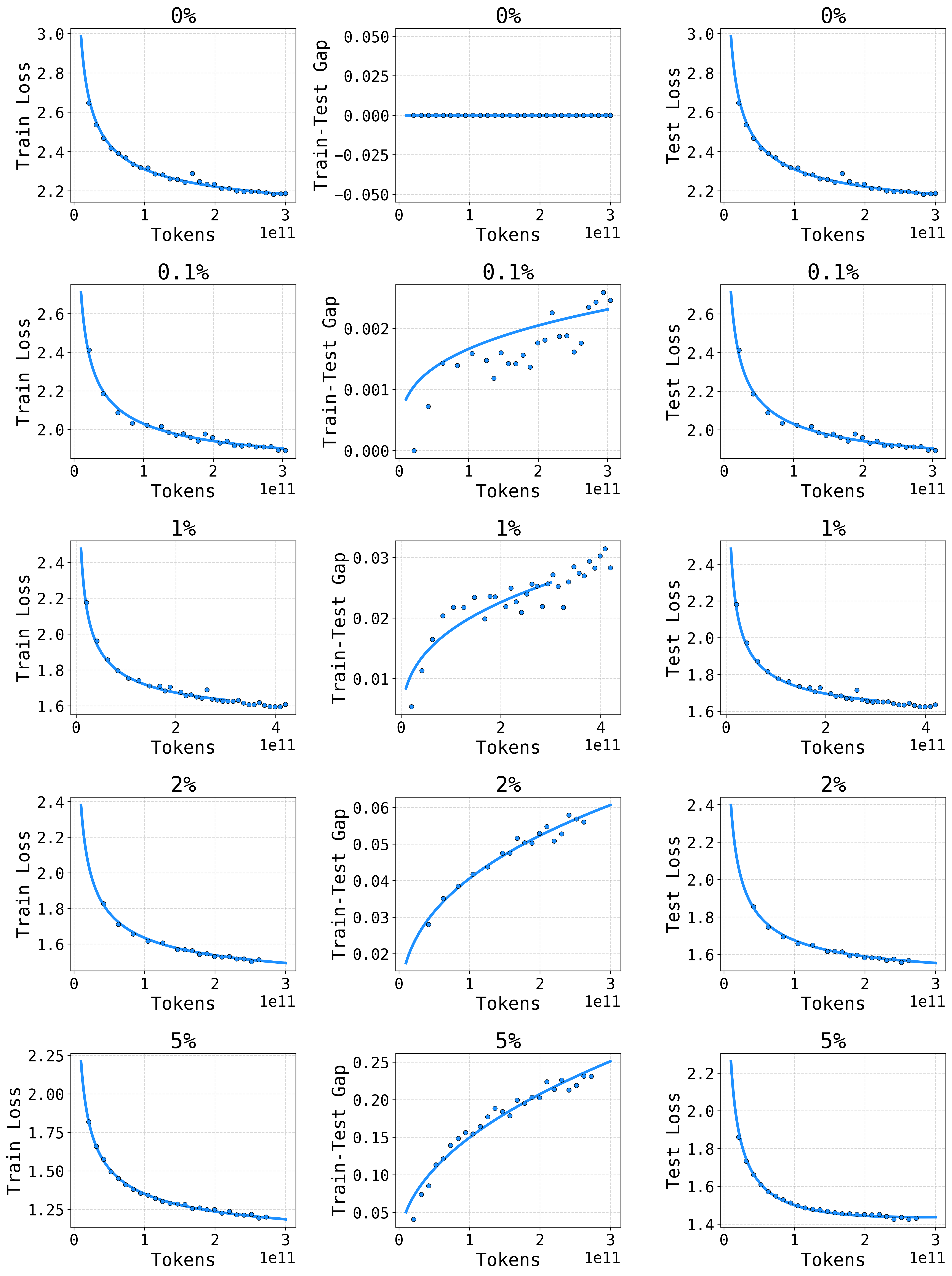}
    \caption{All power law fits over the training loss, train-test gap on \texttt{ChemPile-300M} over the course of SPT. Finally, test loss as the sum of two powers. \label{fig:chempile_separate_fits}}
\end{figure}

\begin{figure}
    \centering
    \includegraphics[width=0.9\textwidth]{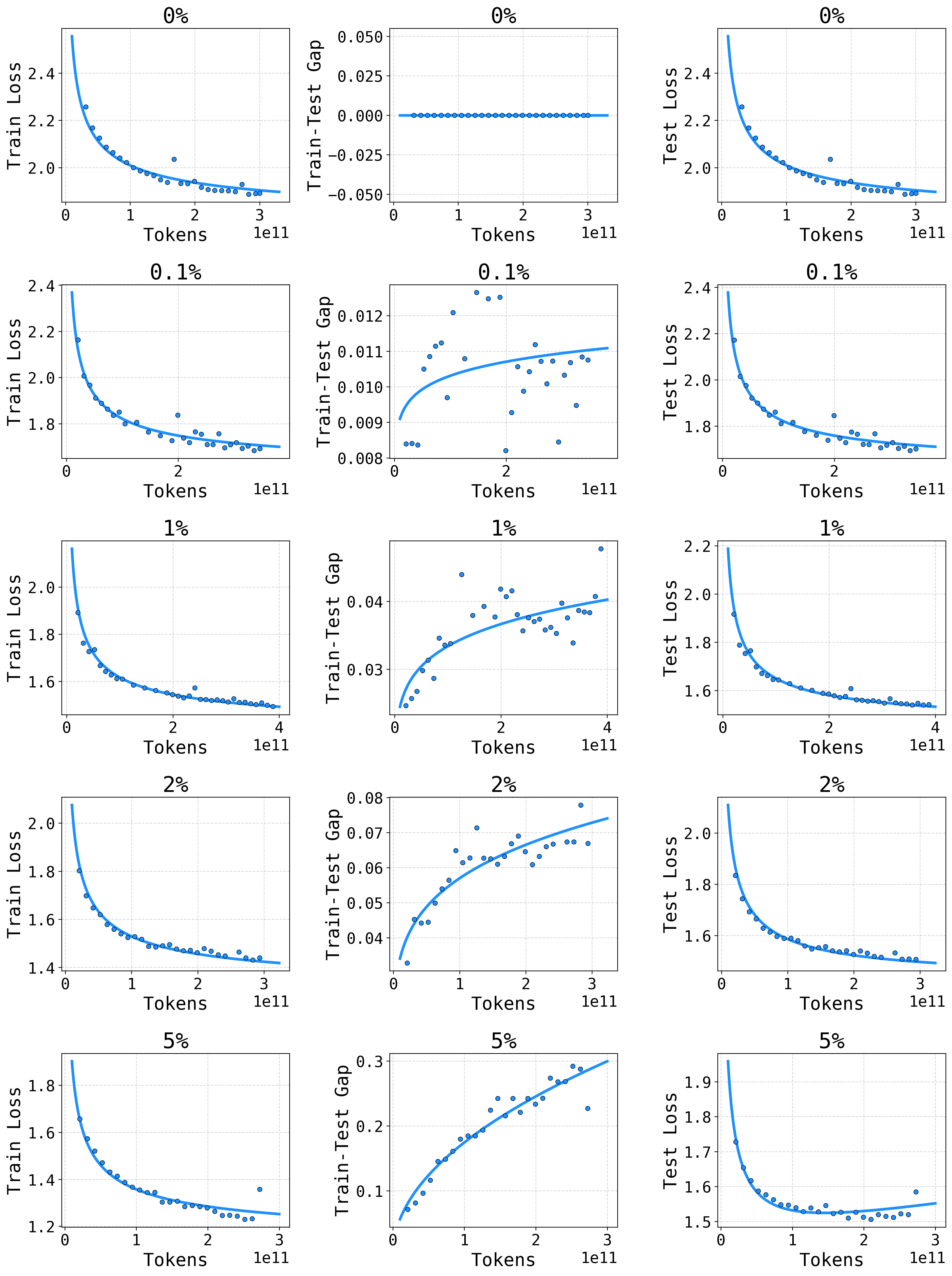}
    \caption{All power law fits over the training loss, train-test gap on \texttt{ProofPile-300M} over the course of SPT. Finally, test loss as the sum of two powers. \label{fig:proofpile_separate_fits}}
\end{figure}

These function choices were made by first learning separate coefficients $A_\delta$ and $C_\delta$ to fit power law curves for each SPT run separately, then identifying the right expression that correctly models these coefficients as a function of $\delta$. Using scipy optimization package, we find power law fits by minimizing the mean squared error over all coefficients $A_\delta, C_\delta \forall \delta \in \{0, 0.1, 1, 2, 5\}$ along with the scalar $A_{\mathrm{train}}$ and the additional two parameters in $b_{\mathrm{x}}(\delta)$.
\begin{align}
\mathcal{L}_{\mathrm{train}}(T,\delta)
&= A_{\mathrm{train}}
   \,T^{\,b_{\mathrm{train}}(\delta)}
 + C_{\delta},\\
\mathcal{L}_{\mathrm{gap}}(T,\delta)
&= A_{\delta}
   \,T^{\,b_{\mathrm{gap}}(\delta)}
\end{align}
Below, we plot the learned coefficients for each $\delta$ as scatter points and demonstrate that they roughly follow the Gamma kernel function for $A_\delta$ and log linear function for $C_\delta$. In Figures \ref{fig:musicpile_separate_fits}, \ref{fig:chempile_separate_fits}, and \ref{fig:proofpile_separate_fits}, we provide the final scaling law fits, where we directly optimize over our final model from Equation \ref{eq:scaling_laws}, jointly learning the $A(\delta)$ and $C(\delta)$ parameters. 

\begin{figure}[h!]
\centering
\includegraphics[width=0.8\textwidth]{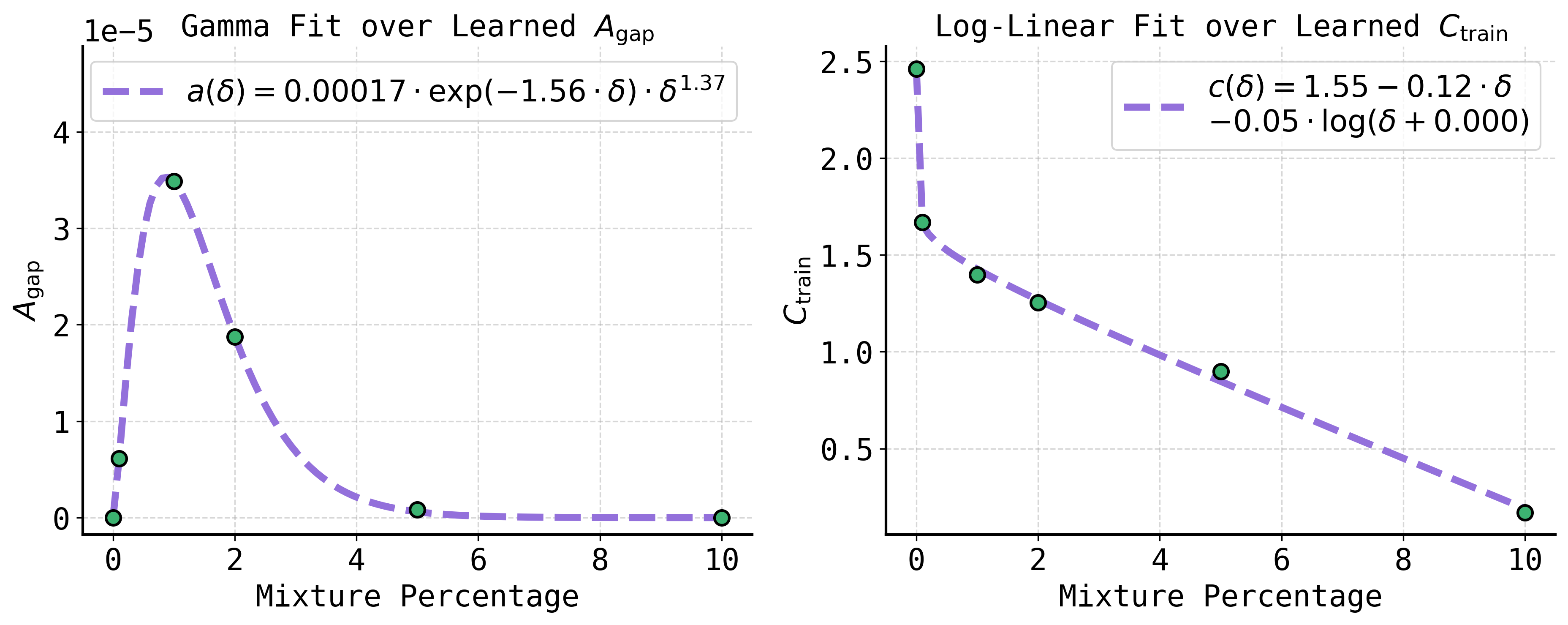}
\includegraphics[width=0.8\textwidth]{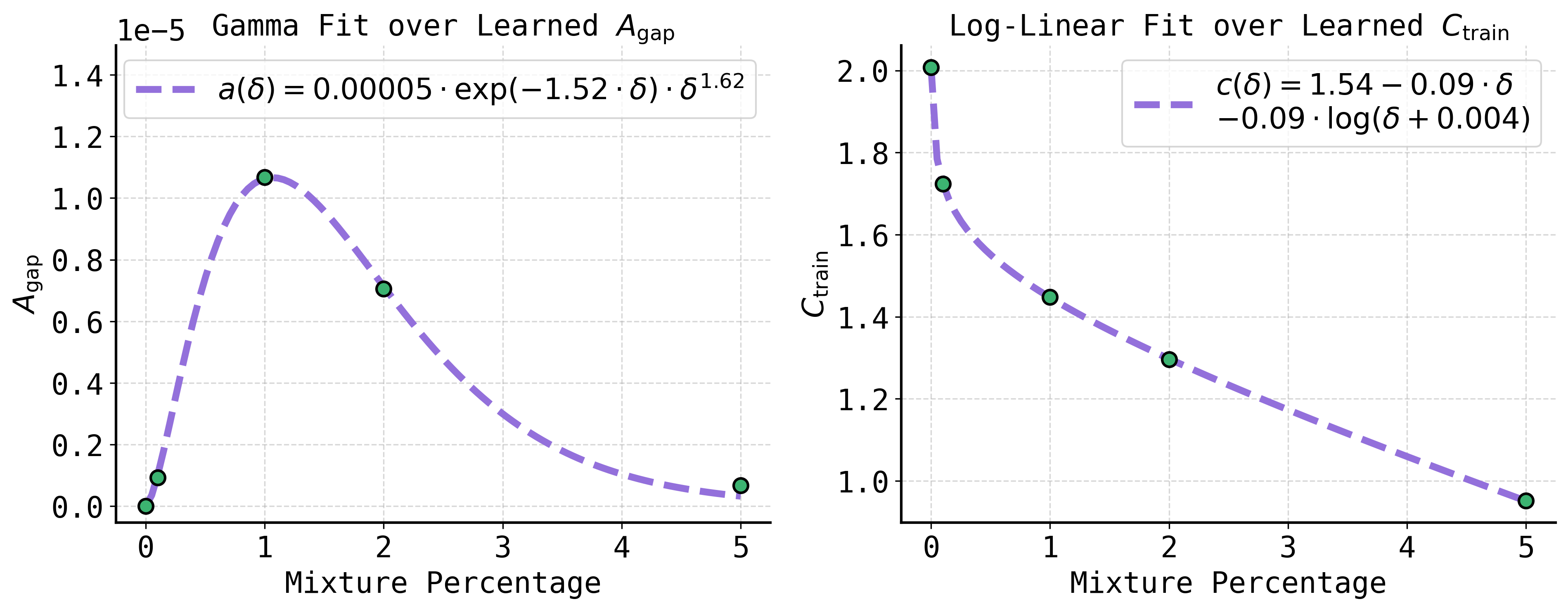}
\includegraphics[width=0.8\textwidth]{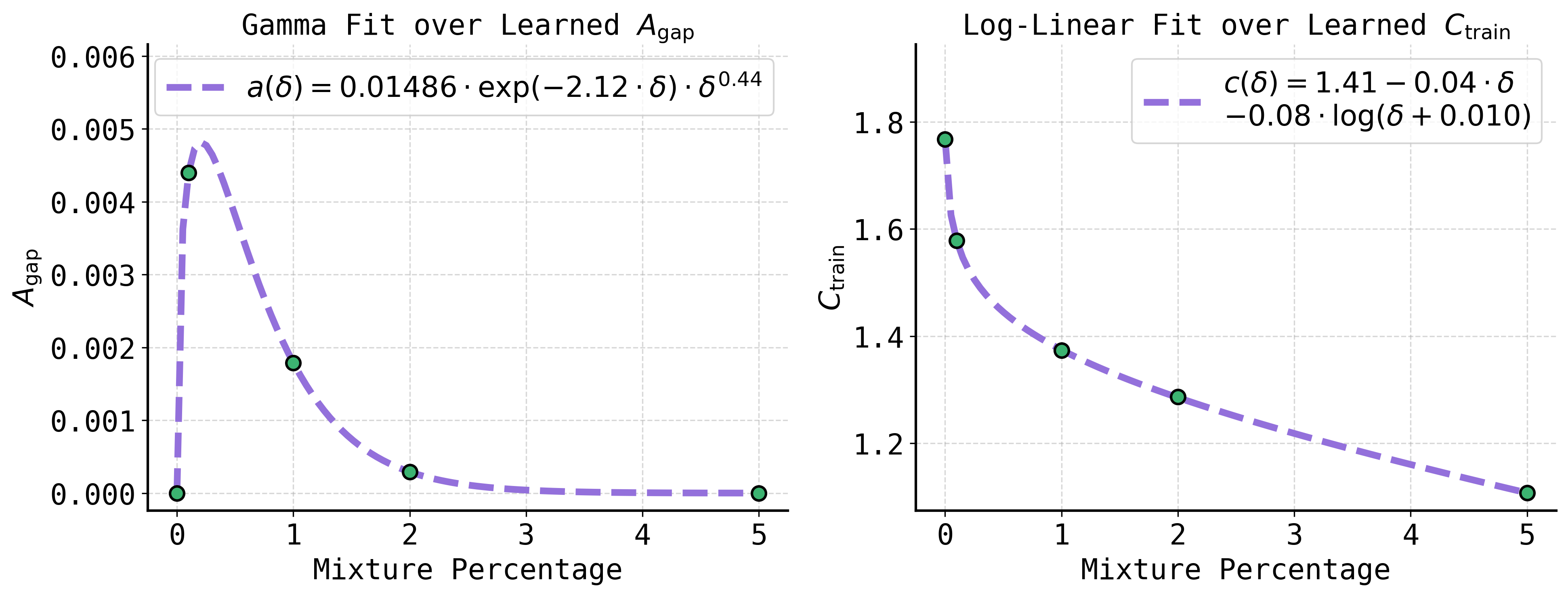}
\caption{Gamma and log-linear fits over learned coefficients.}
\end{figure}

\FloatBarrier

\subsection{Modeling Difference in Test Loss Post-Finetuning Across PT Tokens}

\begin{figure}[h]
\centering 

\begin{subfigure}[b]{\textwidth}
\centering 

\includegraphics[width=0.57\textwidth]{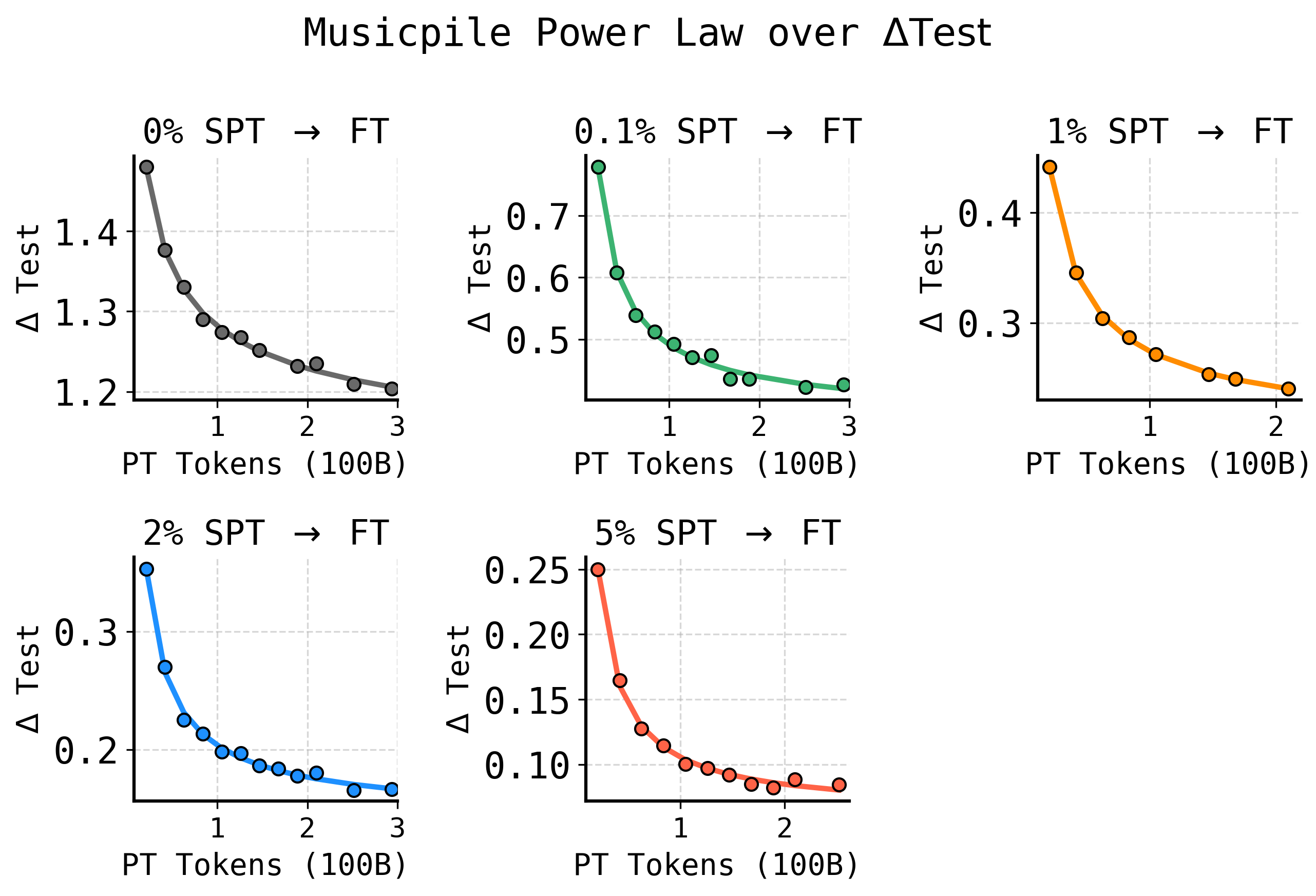}
\caption{MusicPile}
\end{subfigure} 

\begin{subfigure}[b]{\textwidth}
\centering 

\includegraphics[width=0.57\textwidth]{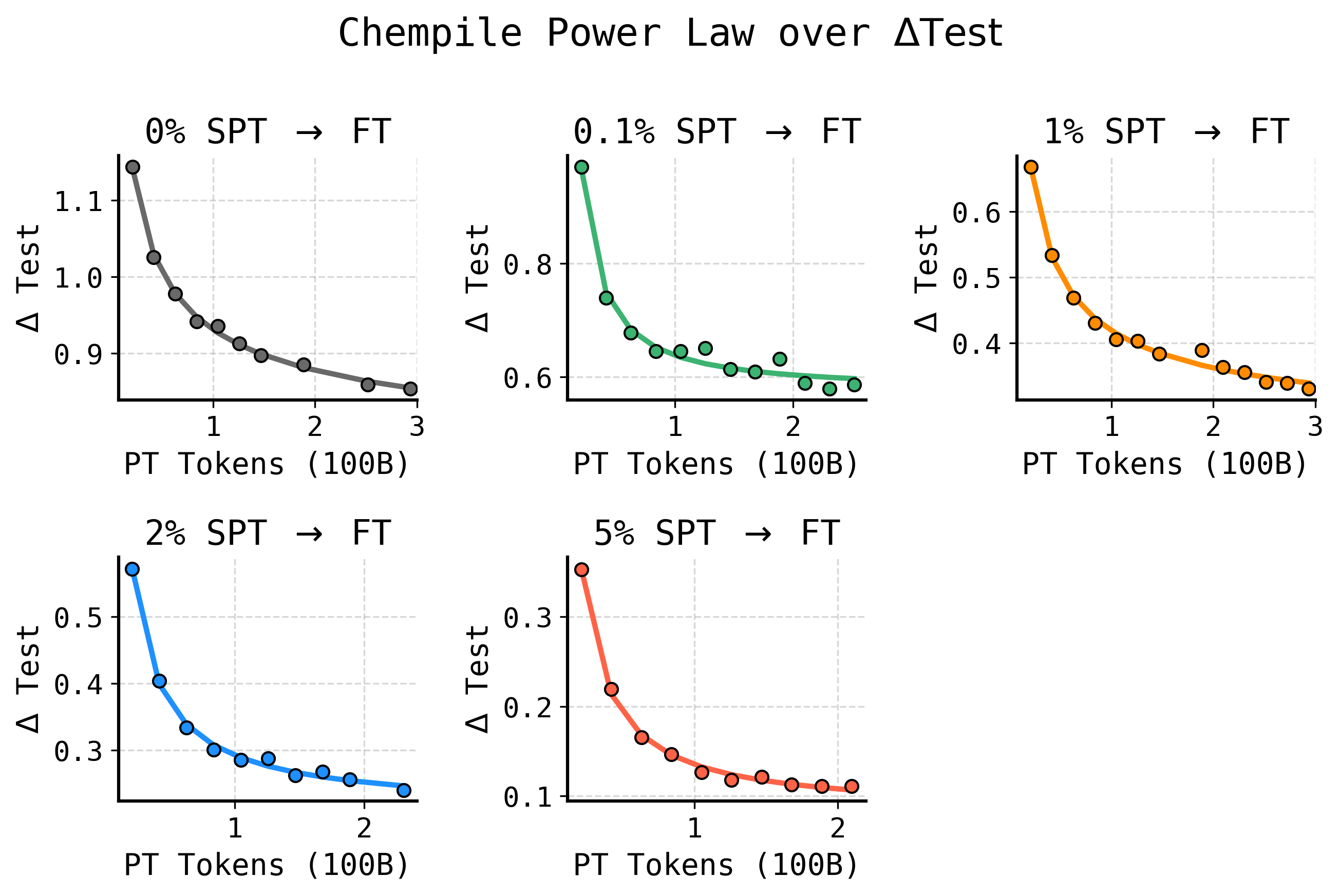}
\caption{ChemPile}
\end{subfigure} 

\begin{subfigure}[b]{\textwidth}
\centering 

\includegraphics[width=0.57\textwidth]{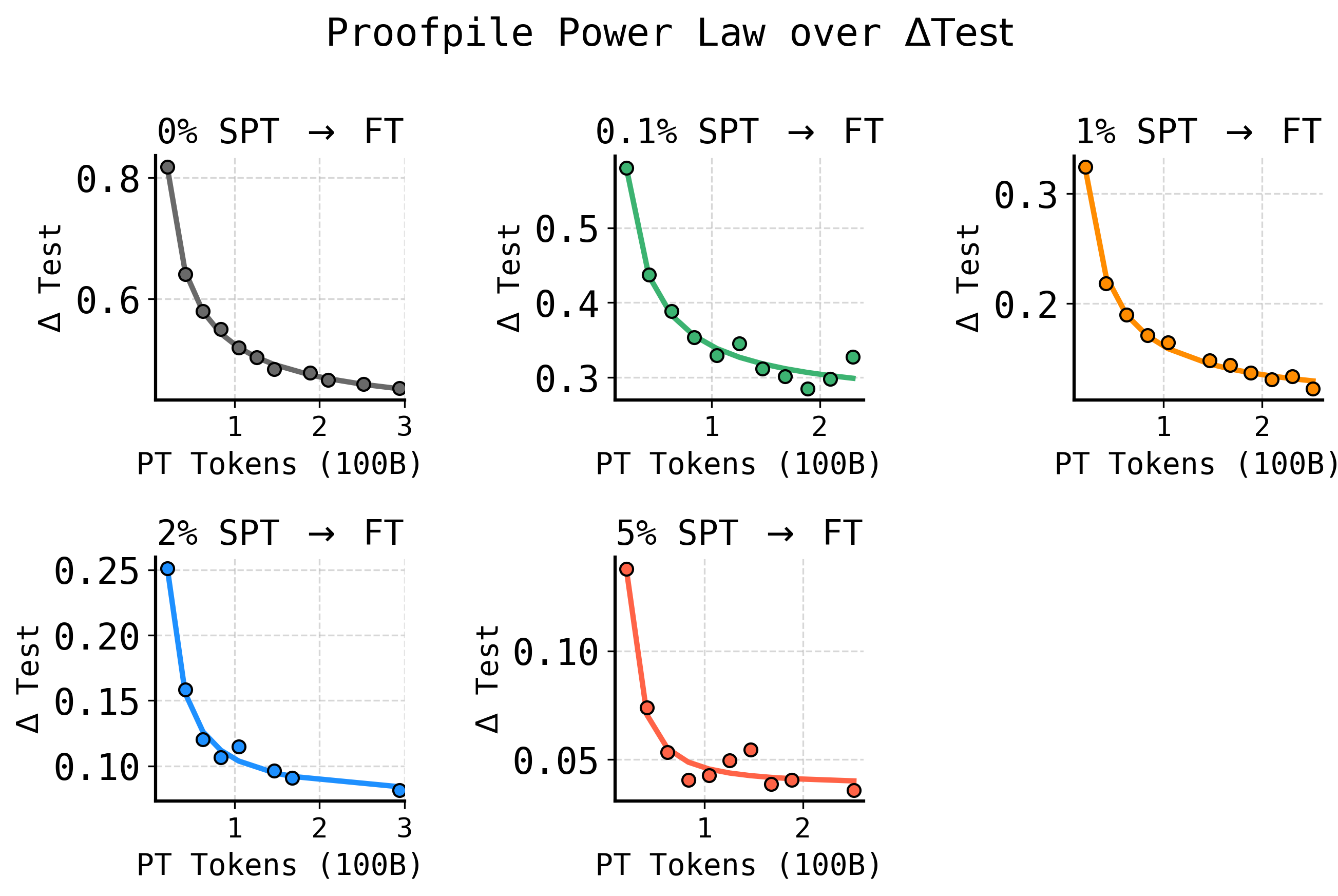}
\caption{ProofPile}
\end{subfigure} 

\caption{$\Delta$ Test follows a power law relationship with respect to the pretraining steps.}
\end{figure}

\FloatBarrier

\newpage
\section{Distribution Similarity Metrics: Detailed Analysis}
\label{app:distribution-metrics}

\begin{figure}[h]
    \centering
    \includegraphics[width=0.9\linewidth]{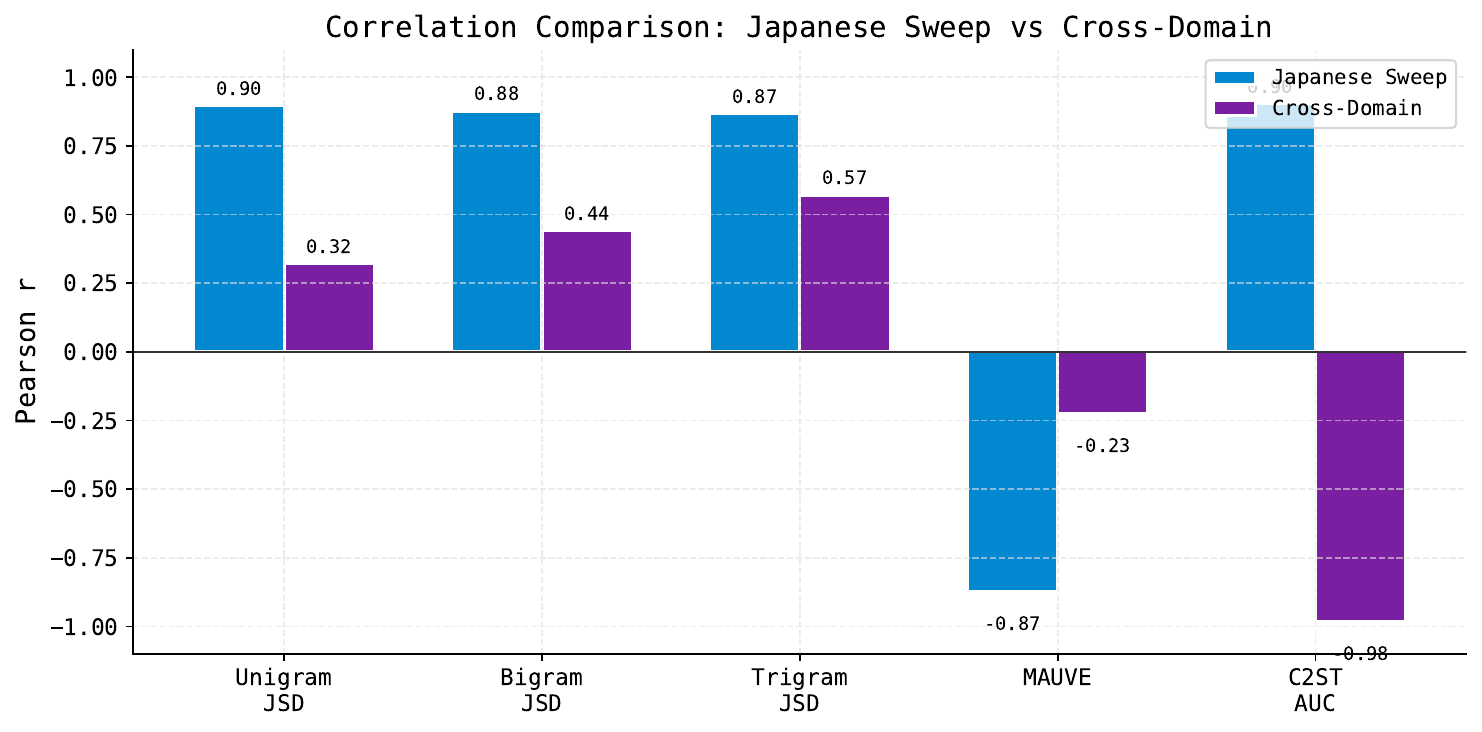}
    \caption{Comparison of Pearson correlations between the Japanese overlap sweep and cross-domain analysis. C2ST AUC flips from $r = +0.90$ to $r = -0.98$.}
    \label{fig:correlation-comparison}
\end{figure}

This appendix describes the distributional similarity metrics used in Section~\ref{sec:domain-similarity} and analyzes how they correlate with $\mathcal{R}_{\text{gain}}$ across our two experimental settings.

\subsection{Metrics Description}

We evaluate five metrics that capture different aspects of how a domain dataset differs from a reference corpus.

\paragraph{Jensen-Shannon Divergence (JSD).}
JSD measures divergence between two probability distributions, bounded between 0 (identical) and 1 bit (maximally different). We compute JSD over n-gram distributions at three granularities: unigram (individual tokens, capturing vocabulary differences), bigram (consecutive pairs, capturing local syntax), and trigram (triplets, capturing longer-range patterns). Higher JSD indicates greater distributional difference. For bigram and trigram computation, we use feature hashing with $2^{20}$ bins to handle the large vocabulary.

\paragraph{MAUVE.}
MAUVE \citep{pillutla2021mauve} compares text distributions using neural embeddings from GPT-2 Large. It computes the area under a divergence frontier curve, producing a score between 0 (completely different) and 1 (identical). We use 5,000 text segments of 256 tokens each per distribution.

\paragraph{Classifier Two-Sample Test (C2ST).}
C2ST~\citep{lopezpaz2016c2st} trains a binary classifier to distinguish samples from two distributions. We use sentence embeddings from all-MiniLM-L6-v2~\citep{wang2020minilm} and train a logistic regression classifier with 5-fold cross-validation. The AUC score indicates separability: 0.5 means indistinguishable, 1.0 means perfectly separable.

\subsection{Experimental Setup}

\paragraph{Japanese Overlap Sweep.}
For the controlled study, we compare mixed pretraining distributions against a held-out English-Japanese parallel translation corpus. The pretraining mixes contain varying percentages of Japanese monolingual web text (0\%, 0.001\%, 0.01\%, 0.1\%, 1\%, and 10\%), with the remainder being English web text. As Japanese percentage increases, the pretraining distribution becomes more similar to the translation target domain. We sample 1.5M tokens from each distribution and compute all metrics against the translation corpus.

\paragraph{Cross-Domain Analysis.}
For the cross-domain analysis, we compare each specialized domain (ChemPile, MusicPile, ProofPile) against the Dolma web corpus used for pretraining. Each domain represents a naturally occurring distributional shift from web text: chemistry literature, symbolic music notation in ABC format, and formal mathematical proofs.

\subsection{Results}

\begin{figure}[h]
    \centering
    \includegraphics[width=\linewidth]{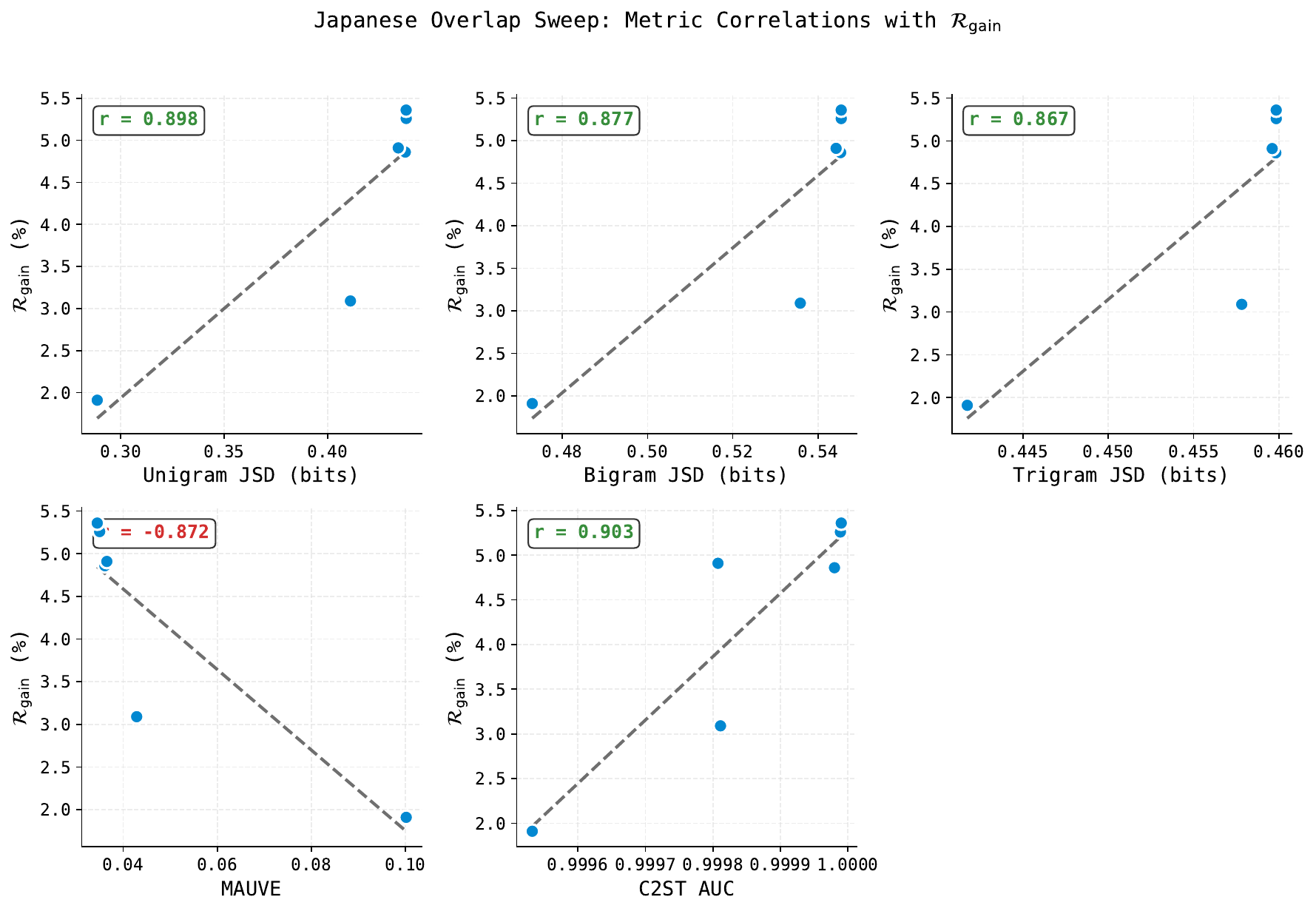}
    \caption{Japanese overlap sweep: each metric plotted against $\mathcal{R}_{\text{gain}}$. All metrics show strong correlations ($|r| > 0.85$) in the expected direction.}
    \label{fig:jp-correlations}
\end{figure}

\paragraph{Japanese Overlap Sweep.}
Figure~\ref{fig:jp-correlations} shows scatter plots of each metric against $\mathcal{R}_{\text{gain}}$ for the Japanese overlap sweep. All five metrics correlate strongly with $\mathcal{R}_{\text{gain}}$ in the expected direction, with Pearson $|r| > 0.85$ in every case. The JSD metrics show positive correlations (higher divergence corresponds to higher gain), MAUVE shows negative correlation (lower similarity corresponds to higher gain), and C2ST AUC shows positive correlation (higher separability corresponds to higher gain). These strong correlations validate that the metrics capture meaningful distributional differences relevant to SPT benefit.

\begin{figure}[h]
    \centering
    \includegraphics[width=\linewidth]{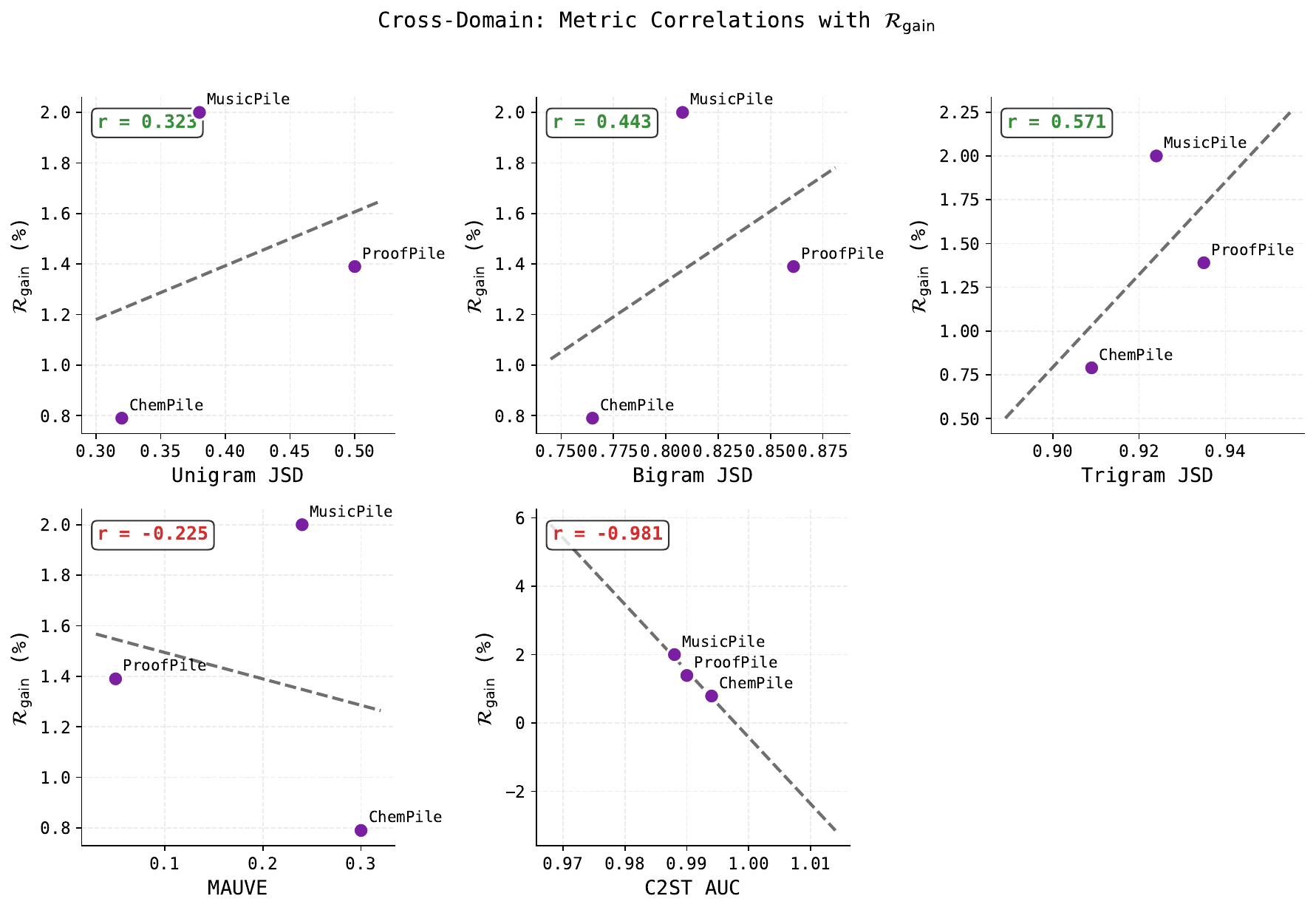}
    \caption{Cross-domain analysis: each metric plotted against $\mathcal{R}_{\text{gain}}$ for ChemPile, MusicPile, and ProofPile. Correlations are much weaker than the Japanese sweep, and C2ST AUC flips sign.}
    \label{fig:cross-correlations}
\end{figure}

\paragraph{Cross-Domain Analysis.}
Figure~\ref{fig:cross-correlations} shows the same analysis for our three benchmark domains. The pattern is strikingly different. JSD metrics show weak positive correlations ranging from $r = 0.32$ (unigram) to $r = 0.57$ (trigram), with longer n-grams performing better. MAUVE shows weak negative correlation ($r = -0.23$). Most dramatically, C2ST AUC flips from $r = +0.90$ on the Japanese sweep to $r = -0.98$ on cross-domain, meaning it predicts the opposite of what we observe.

\paragraph{Why Does C2ST Flip?}
On the Japanese sweep, C2ST correctly identifies that pretraining mixes with more Japanese text are more similar to the translation target, and this similarity predicts lower $\mathcal{R}_{\text{gain}}$. But on cross-domain, C2ST suggests that MusicPile (lowest AUC of 0.988) is most similar to Dolma, when in fact MusicPile shows the highest $\mathcal{R}_{\text{gain}}$. The likely explanation is that C2ST's sentence embeddings capture surface-level semantic similarity that does not generalize across fundamentally different domain types. Music notation in ABC format may appear ``similar'' to web text in embedding space because both contain ASCII characters and structured patterns, even though the underlying content is entirely different.

\section{General Pretraining Loss during SPT}
\label{app:general_loss}
We plot the Dolma loss over pretraining scales across all domains and SPT configurations. As we discuss in Section \ref{subsec:downstream}, SPT replaces a small fraction of general pretraining tokens with domain-specific data, but this has a marginal impact on the general pretraining loss over Dolma for mixture percentages up to $5\%$. However, we do observe notable degradation once we push the percentage up to $10\%$. 

\begin{figure}[h!]
\includegraphics[width=\textwidth]{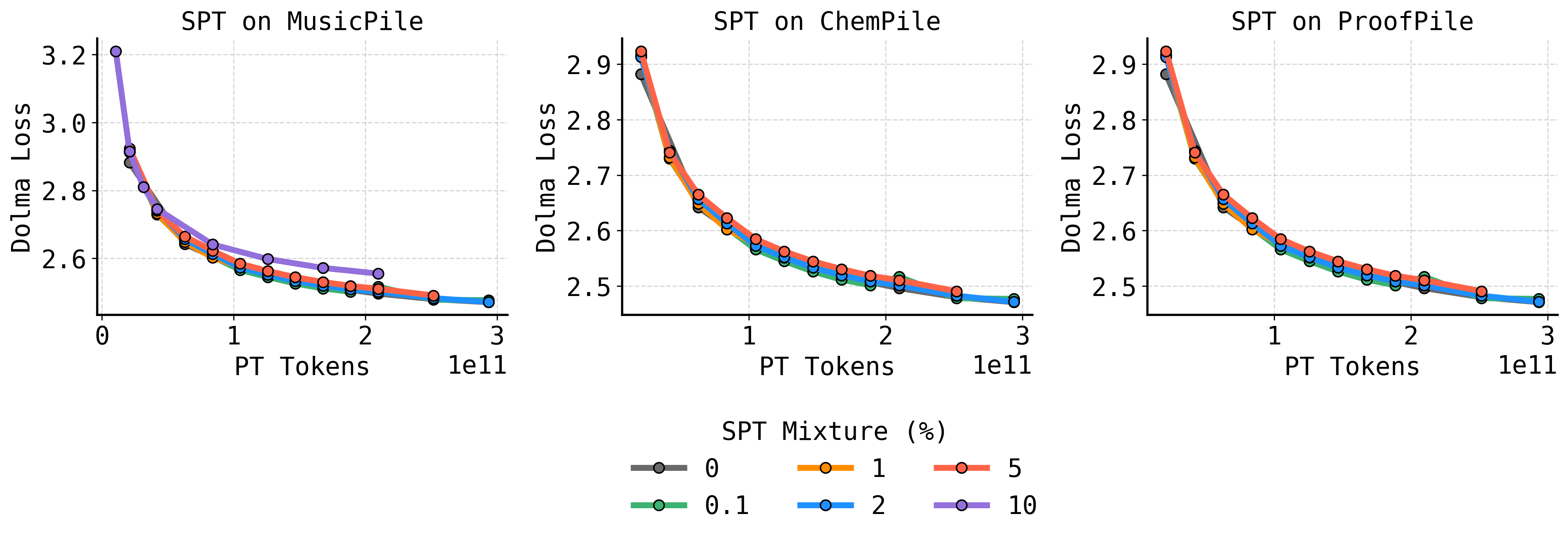}
\end{figure}

\end{document}